\pdfoutput=1

\documentclass[11pt]{article}

\usepackage{acl}

\usepackage{times}
\usepackage{latexsym}

\usepackage[T1]{fontenc}

\usepackage[utf8]{inputenc}

\usepackage{microtype}

\usepackage{inconsolata}

\usepackage[utf8]{inputenc} %
\usepackage[T1]{fontenc}    %
\usepackage{hyperref}       %
\usepackage{url}            %
\usepackage{xspace}
\usepackage{booktabs}       %
\usepackage{amsfonts}       %
\usepackage{nicefrac}       %
\usepackage{microtype}      %
\usepackage{xcolor}         %

\usepackage{amsmath}
\usepackage{bm}
\usepackage{mathtools}
\usepackage{amsthm}
\usepackage{algorithm}
\usepackage{algorithmic}
\usepackage{wrapfig}
\usepackage{enumitem}
\usepackage{makecell}

\usepackage{graphicx} %
\usepackage{subcaption}
\usepackage[font=small]{caption}
\usepackage{multicol}
\usepackage{multirow}
\usepackage{booktabs}
\usepackage{colortbl}
\usepackage[most]{tcolorbox}

\title{Learning Personalized Alignment in Evaluating Open-ended Text Generation}

\author{Danqing Wang$^{1,2}$ \quad Kevin Yang$^{2}$ \quad Hanlin Zhu$^{2,3}$ \quad  Xiaomeng Yang$^{2}$  \\  
\textbf{Andrew Cohen$^{2}$ \quad Lei Li$^{1}$ \quad Yuandong Tian$^{2}$} \\
$^1$Carnegie Mellon University
\quad $^2$Meta AI 
\quad $^3$ UC Berkeley\\
\texttt{danqingw@andrew.cmu.edu} \\ 
\texttt{\{yangkevin,andrewcohen,yuandong\}@meta.com} \\
\texttt{bit.yangxm@gmail.com} \quad
\texttt{hanlinzhu@berkeley.edu} \quad
\texttt{leili@cs.cmu.edu}
}

\newcommand{\method}{\textbf{\textsc{PerSE}}\xspace}
\newcommand{\dataseta}{Per-MPST\xspace}
\newcommand{\datasetb}{Per-DOC\xspace}

\begin{document}

\maketitle

\newcommand{\yuandong}[1]{\textcolor{red}{[Yuandong: #1]}}
\newcommand{\kevin}[1]{\textcolor{cyan}{[Kevin: #1]}}
\newcommand{\revise}[1]{\textcolor{red}{#1}}

\begin{abstract}
\label{sec:abs}

Recent research has increasingly focused on evaluating large language models' (LLMs) alignment with diverse human values and preferences, particularly for open-ended tasks like story generation. Traditional evaluation metrics rely heavily on lexical similarity with human-written references, often showing poor correlation with human judgments and failing to account for alignment with the diversity of human preferences. To address these challenges, we introduce \method, an interpretable evaluation framework designed to assess alignment with specific human preferences. It is tuned to infer specific preferences from an in-context personal profile and evaluate the alignment between the generated content and personal preferences. \method enhances interpretability by providing detailed comments and fine-grained scoring, facilitating more personalized content generation. Our 13B LLaMA-2-based \method shows a 15.8\% increase in Kendall correlation and a 13.7\% rise in accuracy with zero-shot reviewers compared to GPT-4. It also outperforms GPT-4 by 46.01\% in Kendall correlation on new domains, indicating its transferability~\footnote{Both datasets and code are released at https://github.com/facebookresearch/perse.}.

\end{abstract}

\section{Introduction}
\label{sec:intro}

\begin{figure}[ht]
    \centering
    \includegraphics[width=0.95\linewidth]{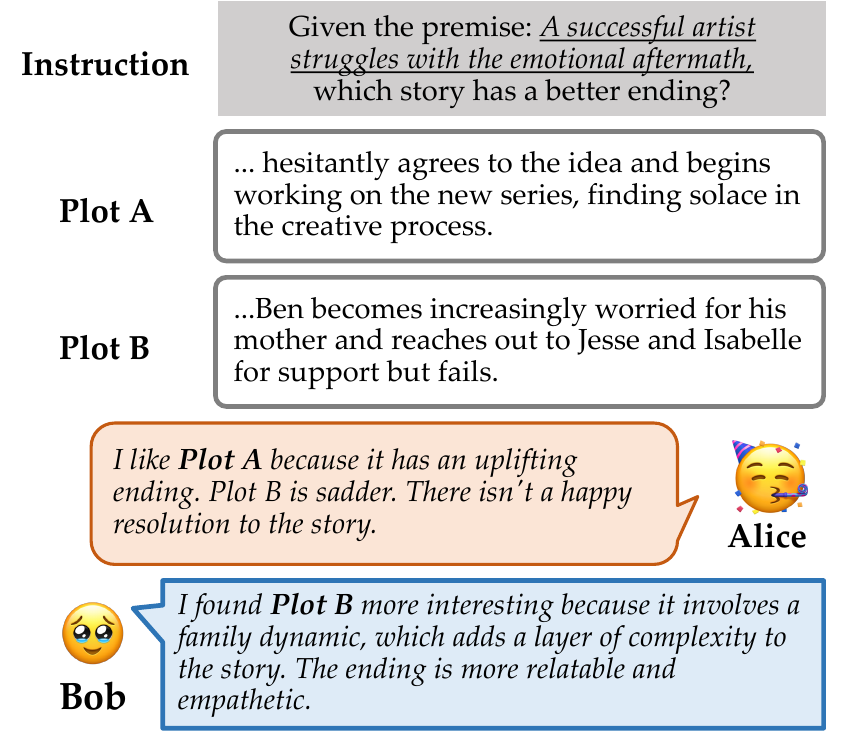}
    \caption{Two human reviewers have distinct preferences for LLM-generated stories from the same premise.}
    \label{fig:personalization}
\end{figure}

Large language models (LLMs) have recently demonstrated impressive generative capabilities across various tasks, with rapid improvements in language qualities such as fluency and consistency~\citep{ouyang2022training,bai2022constitutional,touvron2023llama2}. However, evaluating their performance in open-ended generation tasks remains challenging due to the diversity of responses. Traditional automatic metrics struggle with the one-to-many problem in open-ended generation~\citep{liu-etal-2016-evaluate} and often show poor correlation with human judgment~\citep{krishna-etal-2021-hurdles,guan2021openmeva}. Recent studies have trained evaluation metrics based on human ratings to better approximate human judgments~\cite{sellam-etal-2020-bleurt,rei-etal-2020-comet}. However, these metrics primarily focus on objective qualities and tend to overlook subjective assessments, such as surprise~\citep{chhun2022human} or interestingness~\citep{bae2021preliminary}.

Subjective evaluation metrics are significantly influenced by diverse human preferences. For example, Figure \ref{fig:personalization} illustrates two stories generated by ~\citet{yang-etal-2023-doc} from the same premise. Alice prefers Plot A for its uplifting ending, while Bob favors Plot B due to its plot complexity and empathetic ending. This highlights the need for an automatic personalized evaluation metric that can assess model generations based on varying preferences. However, it is costly for each reviewer to provide a large number of personalized examples to demonstrate their preferences. This makes it impractical to train a separate evaluation model for each reviewer and to generalize the existing metric to unseen reviewers.

Moreover, the subjective nature of these evaluations makes the scores harder to interpret. AuPEL~\citep{wang2023automated} incorporates personalization as one of the evaluation aspects to compare two inputs, but it does so without any explanation. This lack of transparency undermines the trustworthiness and reliability of evaluations and complicates the development of generative models~\citep{leiter2022towards}. Therefore, the key challenges in personalized evaluation are modeling an unseen reviewer's preference from a limited annotated personalized context and providing an interpretable explanation for the assessment.

In this paper, we introduce an LLM-based evaluation model, \method, designed to assess the alignment between open-ended generations and specific preferences. \method is tuned to infer preferences from a limited-length profile and uses this information to evaluate the generated content. \method provides an overall score along with an explanation for the scalar rating and offers fine-grained scores on several aspects to interpret the alignment for pairwise ratings. We curated two instruction-following datasets, \dataseta and \datasetb, to support personalized alignment in evaluation. \method is fine-tuned from LLaMA-2~\citep{touvron2023llama2} to enhance its capability to infer preferences from reviewer profiles and apply these preferences in evaluations. Compared with GPT-4, \method achieves a 15.8\% higher Kendall correlation in the scalar rating of movie plot generation and a 13.7\% higher accuracy in the pairwise rating of story generation for zero-shot reviewers. It also outperforms GPT-4 by 46.01\% in Kendall correlation when transferred to new domains. Our contributions can be summarized as follows:

\begin{itemize}[noitemsep, leftmargin=1em, topsep=1pt]
    \item We develop an LLM-based evaluation model, \method, to assess alignment between open-ended generations and in-context preferences. By instruction-tuning on personalized data, it significantly outperforms GPT-4 in evaluating personal alignment.
    \item \method provides detailed explanations for its assessments. Its interpretability makes it particularly suitable for guiding personalized content generation.
    \item We curate two instruction-following datasets specifically for personalized alignment in the evaluation of open-ended generations.
    \item We find that LLMs, after reinforcement learning via human feedback, tend to be less personalized and more cautious with negative comments, which hinders their ability to align with strong personal preferences. However, when instruction-tuned with personalized data, even less powerful LLMs can perform better in aligning with preferences.

\end{itemize}

\section{Related Work}
\label{sec:related}
\begin{figure*}[ht]
    \centering
    \includegraphics[width=1\linewidth]{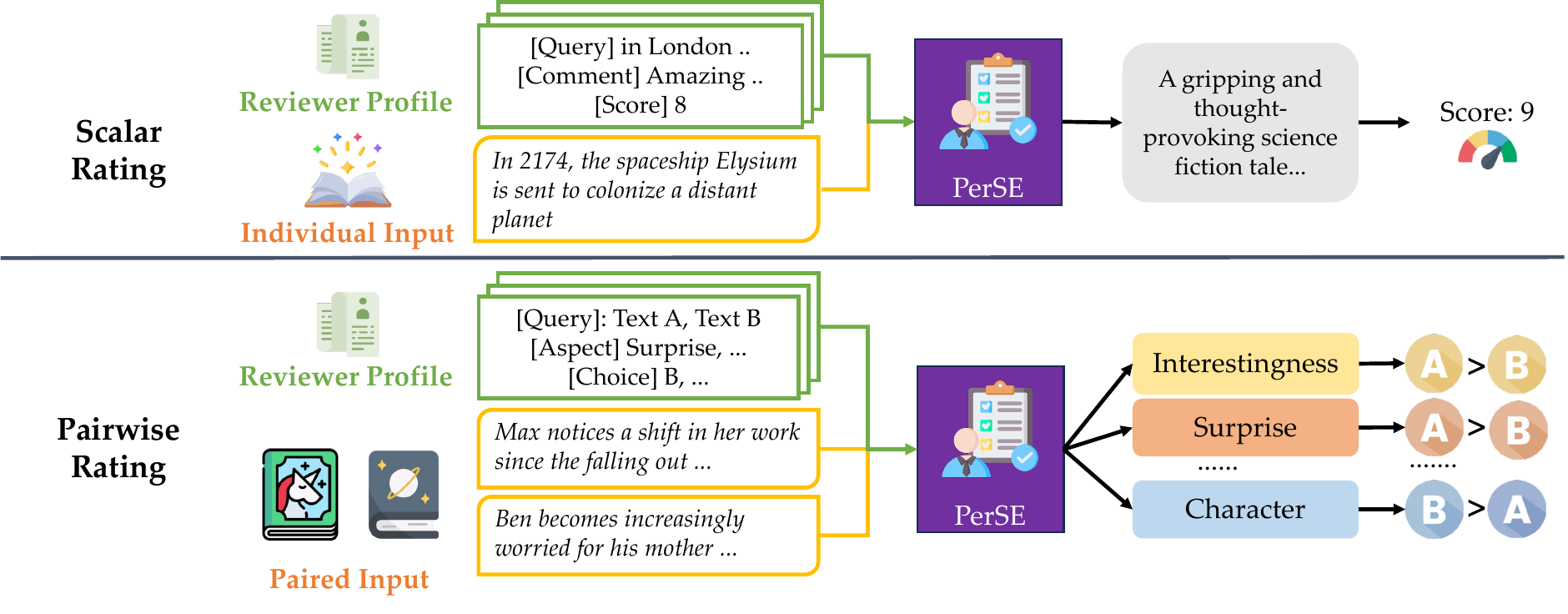}
    \caption{\method provides the scalar rating and pairwise rating for the personalized alignment in evaluation. \method infers the reviewer's preference from the profile with historical reviews and employs the preference to provide an interpretable evaluation.}
    \label{fig:overview}
\end{figure*}

\textbf{Evaluation Metrics for Text Generation}
Automatic metrics can be broadly categorized into reference-based and reference-free metrics. Reference-based metrics evaluate the similarity between the reference and the model output based on lexical overlap~\citep{papineni-etal-2002-bleu,lin-2004-rouge} or embedding distance~\citep{zhang2019bertscore,zhao2019moverscore}. In contrast, reference-free metrics directly assess the quality of the model output without any reference. These metrics are usually trained to evaluate generation from an overall perspective~\citep{guan-huang-2020-union,ghazarian-etal-2021-plot} or along multiple axes~\citep{chen2022storyer,xie2023deltascore}. Recently, researchers have explored using large language models in evaluation metrics~\citep{fu2023gptscore,kocmi2023large,xu-etal-2023-instructscore}, or as judges~\citep{Bai2023BenchmarkingFM,zheng2023judging}. In this paper, we investigate LLMs' capabilities in learning personalized alignment on subjective aspects, which is crucial for evaluating open-ended generation.

\noindent \textbf{Human Evaluation for Generation}
Human evaluation is employed to assess various aspects of text quality, such as coherence~\citep{xu-etal-2018-skeleton,peng-etal-2018-towards}, relevance~\citep{yang-etal-2023-doc,yang2022re3,jhamtani-berg-kirkpatrick-2020-narrative}, and interestingness~\citep{bae2021preliminary}. To comprehensively cover all aspects, \citet{chhun2022human} suggested six human criteria for storytelling: relevance, coherence, empathy, surprise, engagement, and complexity. However, they found that the inter-annotator agreement for human evaluation on these subjective aspects is low. \citet{karpinska-etal-2021-perils} also pointed out the risks of crowdsourced human judgments from Amazon Mechanical Turk due to under-qualified workers and a lack of reproducibility details.

\noindent \textbf{Personalization in Text Generation and Evaluation}
Personalization has been extensively studied in many recommendation systems~\citep{das2007google,xu2022rethinking} and search applications~\citep{croft2001relevance,shi2022every}. Recently, researchers have emphasized its importance in natural language processing~\citep{flek-2020-returning,dudy-etal-2021-refocusing}. Several recent studies have explored LLMs' capabilities in capturing personalization~\citep{chen2023large,kang2023llms,salemi2023lamp} or in prompting for personalized recommendations~\citep{lyu2023llm,chen2023palr,li2023prompt}. \citet{wang2023automated} introduced a personalization score as one of the evaluation aspects, using LLMs as evaluators. In this paper, we propose an interpretable evaluation model to align with personal preferences, which not only provides an assessment but also a detailed explanation.

\section{Learning Personalized Alignment in Evaluating Open-ended Generation}
\label{sec:method}
We propose an LLM-based evaluation model for assessing the alignment between generated content and personal preferences. \method provides a reference-free, interpretable evaluation from a specific reviewer's perspective.

Given an open-ended generation as the query $\bm{x}$ and the personal preference $p_u$ of reviewer $u$, the goal is to provide a scalar or pairwise rating $y_u$ along with an explanation $\bm{e}_u$ to indicate how well the query aligns with the reviewer's personal preference. The reviewer's preference is defined by their profile, which includes their historical comments $p_u = \{(\bm{x}^{(i)}, \bm{e}_u^{(i)}, y_u^{(i)})\}_{i=0}^{K}$, where $K$ is the number of historical comments. The tuple $(\bm{x}^{(i)}, \bm{e}_u^{(i)}, y_u^{(i)})$ is sampled from the historical set $\mathcal{M}_p$. Typically, the number of a reviewer's historical comments is limited, making it challenging to train a separate rating model for each reviewer\footnote{For simplicity, we assume the reviewer's preferences are consistent within the review time frame.}.

As demonstrated in Figure \ref{fig:overview}, \method provides a scalar rating for individual input text and a pairwise rating between two inputs. For the scalar rating, the reviewer's comment serves as the explanation $\bm{e}_u$, and their rating (ranging from 1 to 10) represents the alignment score $y_u$. For a more interpretable comparison between two inputs, we conduct a fine-grained assessment across a set of aspects $\mathcal{A}$. For each aspect, \method determines the winner between the two inputs and uses this as the rating. The aspects and their ratings collectively form the explanation $\bm{e}_u$. The win rate across all aspects is considered the alignment score $y_u$.

\subsection{Evaluate with Personal Preference}
\label{sec:eval}
We use LLMs to evaluate open-ended generations based on the reviewer's profile $p_u$. The LLM is instructed to analyze the implicit personal preferences from the historical comments in the profile $p_u$ and predict the score $y_u$ and explanation $\bm{e}_u$ for the new query $\bm{x}$ based on these preferences. In scalar ratings, the query $\bm{x}$ is a single generation, while in pairwise ratings, it consists of two generations. A high score indicates that the generation aligns well with the preference and suggests that the reviewer is likely to rank it highly. Conversely, a low score suggests that the reviewer may not favor it. For the same query $\bm{x}$, there could be distinct scores $y_{u_i} \neq y_{u_j}$ for different reviewers $u_i \neq u_j$. The prompt templates are listed in Figure \ref{fig:perse} in the Appendix.

\subsection{Curate Personalized Alignment Datasets}
\label{sec:curate}
To learn the alignment between personal preferences and generated content, we curate personalized alignment data for evaluating open-ended generation. We utilize two data sources: existing evaluation datasets that include reviewer identity and crowd-sourced annotations for open-ended generation systems (such as MTurk).

However, we find that if a dataset has been exposed to most LLMs during their pre-training, it can suffer from severe contamination issues during evaluation. The LLM-based evaluation model may memorize the ground-truth scores and perform well on the exposed test set, but its performance drops dramatically on non-memorized data, making it difficult to generalize to new cases. We discuss the influence of contamination issues on the evaluation of open-ended generation in Appendix \ref{sec:a_mem}.

To address this, we recreate existing evaluation datasets to alleviate contamination issues through anonymization and summarization. We extract identifiable entities, such as characters and locations, and replace them randomly. We then summarize the raw content to omit details while retaining the main content. These strategies help significantly resolve the contamination issue. The details of the data processing and an analysis of its effect are provided in Appendix \ref{sec:a_data_collection}.

\subsection{Learning Personalized Alignment in Evaluation}
\label{sec:ft}
The goal is to enhance LLMs' capabilities in inferring new preferences from an in-context reviewer's profile and applying these preferences in evaluation. We split our dataset into two non-overlapping sets based on the query: the historical reviews $\mathcal{M}_p$ is used for the reviewer profile, and $\mathcal{D}$ is for personalized alignment. We also divide the reviewers in the dataset into $\mathcal{U}_{\rm{ift}}$ and $\mathcal{U}_{\rm{test}}$, which are used for instruction-tuning and testing, respectively. Importantly, there is no overlap between the reviewers in $\mathcal{U}_{\rm{ift}}$ and $\mathcal{U}_{\rm{test}}$, ensuring that the model cannot directly apply any memorized preferences from fine-tuning during inference time.

The instruction-tuning dataset is defined as $D_{\rm{ift}} = \{(\bm{x}, p_u, \bm{e}_u, y_u) | (\bm{x}, \bm{e}_u, y_u) \in \mathcal{D}, u \in \mathcal{U}_{\rm{ift}} \}$, where $p_u \subseteq \{(\bm{x}, \bm{e}_r, y_r) \in \mathcal{M}_p\ | r = u\}$. This dataset includes all reviewers from $\mathcal{U}_{\rm{ift}}$, with profiles built on the historical set $\mathcal{M}_p$. \method learns personalized alignment for evaluation based on instruction fine-tuning on $D_{\rm{ift}}$.

\section{Experiment}
\label{sec:bias}

We introduce our two datasets in Section \ref{sec:dataset}. In Section \ref{sec:baseline}, we describe the implementation of \method and several baselines. Further details on the fine-tuning process are listed in Appendix \ref{sec:training}.

\subsection{Datasets}

\label{sec:dataset}
We recreate our dataset, \dataseta, from the existing movie review dataset MPST~\citep{KAR18.332,kar-etal-2020-multi}, as detailed in Section \ref{sec:curate}. We reorganize the released human annotations from \citet{zhu2023end} for personalized alignment, resulting in \datasetb. As described in Section \ref{sec:ft}, we split the reviewers by a 9:1 ratio into $\mathcal{U}_{\rm{ift}}$ and $\mathcal{U}_{\rm{test}}$, building the instruction-tuning and test sets based on these groups. The instruction-tuning set is used for learning personalized preference alignment, while the test set evaluates model performance on unseen reviewers. The statistics are listed in Table \ref{tab:dataset}.

\textbf{\dataseta} MPST is a movie review dataset collected from IMDb\footnote{https://www.imdb.com/}. It includes a synopsis and multiple comments for each movie. Each comment contains a review text and a score ranging from 1 (lowest) to 10 (highest). We group comments by reviewer ID and remove reviewers with fewer than 6 comments to ensure there are at least 5 historical comments. We create different versions by sampling various numbers of historical reviews ($K=1$ to $5$) as the reviewer profile. Additionally, we remove queries with more than 2500 words (about 4k tokens) to fit within the context window of LLMs.

\textbf{\datasetb} This dataset contains 7,000 unique examples from 403 annotators, based on the released annotations from \citet{zhu2023end}. Each example consists of two plots generated from the same premise, and annotators were asked to answer various questions and choose their preferred plot for each question. 
We define five subjective aspects: \texttt{Interestingness} (I), \texttt{Adaptability} (A), \texttt{Surprise} (S), \texttt{Character Development} (C), and \texttt{Ending} (E). \texttt{Interestingness} focuses on the appeal of the overall narrative; \texttt{Surprise} indicates unexpected elements or twists in the plot; \texttt{Character Development} evaluates the emotional and personal connection between characters and events; \texttt{Ending} pertains to the satisfaction or appreciation of the ending, and \texttt{Adaptability} measures the potential for further developing the story.
We removed annotators with fewer than 2 annotations and use $K=1$ for the reviewer profile.

\begin{table*}[htbp]\footnotesize\setlength{\tabcolsep}{3pt}
  \centering
  \caption{Statistics of \dataseta and \datasetb. Length is the number of words in the instruction, which includes the instruction template, reviewer preference, and plot query. \textbf{I}, \textbf{A}, \textbf{S}, \textbf{C}, and \textbf{E} stand for \texttt{Interestingness}, \texttt{Adaptability}, \texttt{Surprise}, \texttt{Character Development}, and \texttt{Ending}. $k$ is the number of reviews; we fix $k=1$ for \datasetb due to the length.}
    \begin{tabular}{clccccc|ccccc}
    \toprule
          & & \multicolumn{5}{c|}{\textbf{\dataseta}} & \multicolumn{5}{c}{\textbf{\datasetb} ($K=1$)} \\
    \midrule
          & & \textbf{k=1} & \textbf{k=2} & \textbf{k=3}  & \textbf{k=4} & \textbf{k=5} & \textbf{I} & \textbf{A} & \textbf{S} & \textbf{C} & \textbf{E} \\
    \midrule
    \multirow{3}[1]{*}{\textbf{Train}} & \# Reviewers & 1412  & 1394  & 1385  & 1369  & 1336  & 172   & 171   & 156   & 160   & 155 \\
    & \# Example & 13254 & 13940 & 13794 & 13480 & 12041 & 1985  & 1856  & 1722  & 1785  & 1574 \\
    & Avg. Length & 868.9 & 1235.2 & 1600.3 & 1964.0 & 2123.3 & 2410.9 & 2413.7 & 2411.7 & 2409.8 & 2409.6 \\
    \midrule
    \multirow{3}[1]{*}{\textbf{Valid}} & \# Reviewers & 92    & 92    & 92    & 92    & 92    & 18    & 18    & 15    & 18    & 15 \\
    & \# Example & 915   & 920   & 920   & 906   & 833   & 234   & 224   & 161   & 162   & 173 \\
    & Avg. Length & 857.9 & 1237.1 & 1597.2 & 1956.1 & 2108.4 & 2402.9 & 2399.2 & 2408.4 & 2421.4 & 2404.3 \\
    \bottomrule
    \end{tabular}%
  \label{tab:dataset}%
\end{table*}%

\subsection{Experimental Setting}
\label{sec:baseline}
We implement \method based on LLaMA-7b-chat and LLaMA-13b-chat, tuning them on the $D_{\rm{ift}}$ of \dataseta and \datasetb for scalar and pairwise ratings, respectively. In our main experiments, we use $k=3$ for \dataseta and $k=1$ for \datasetb. During inference, we set the temperature to 0.8 and limit the maximum generation length to 600 tokens. We report Pearson, Spearman, and Kendall-Tau correlation coefficients to measure the agreement between ground-truth scores and the predicted scores $y_u$ in scalar ratings. For pairwise ratings, we report the accuracy for each aspect.

\textbf{Baseline} 
We establish a basic baseline that directly uses the average scores from historical reviews as the predicted score. For $k=1$, we use the historical score as the output. This baseline is named \textbf{Reviewer Avg.}, reflecting the average score this reviewer gives based on historical comments. 
On \dataseta, we add the baseline \textbf{Matrix Factorization (MF)}~\citep{koren2009matrix}, commonly used in recommendation systems. It decomposes the user-item interaction matrix into the product of the user matrix and the product matrix, with the main idea being to recommend products based on the similarity between the user and the product. These two baselines do not provide an interpretable explanation for their evaluation.
On \datasetb, both the plot pairs and the annotators in the test set have no overlap with the instruction-tuning set, making the matrix factorization baseline unsuitable in this case. 
We also evaluate the capabilities of vanilla LLMs, including LLaMA-2-chat models from 7b to 70b and GPT-4~\footnote{We used the \texttt{gpt-4-0613} version from \url{https://openai.com/gpt-4} with default settings.}, using the same prompts and generation configurations.

\section{Results and Analysis}
\label{sec:exper}
We report the performance of the scalar rating on the test set of \dataseta and the pairwise rating on \datasetb. The reviewers in the test set $\mathcal{U}_{\rm{test}}$ have no overlap with those in the instruction-tuning set $\mathcal{U}_{\rm{ift}}$.

\subsection{Key Findings}

\noindent \textbf{\method's Scalar Rating Achieves the Highest Correlation with Human Ratings.} As shown in Table \ref{tab:main_movie}, \method-13b significantly outperforms all baselines in terms of correlations with human ratings. Specifically, \method-13b achieves a Pearson correlation of 0.345 between its predictions and human scores, indicating that \method effectively captures the reviewer's preferences from the given profile. The comparison between \method and the vanilla LLMs demonstrates that it is challenging for vanilla LLMs to align evaluations with personal preferences without personalized alignment instruction tuning. 
Moreover, we observe that both the average score of the reviewer's historical reviews and the simple baseline MF are strong baselines. This observation aligns with \citet{kang2023llms}, which shows that vanilla LLMs struggle to understand user preferences. One possible reason is that both the pre-training phase and instruction-tuning via reinforcement learning with human feedback (RLHF) focus on aligning the model towards objective and common human values, which hinders their ability to provide more personalized responses. This is also noted by \citet{kirk2023personalisation}, who claim that the aggregate fine-tuning process may not adequately represent all human preferences and values. However, \method demonstrates that this capability can be easily regained through instruction-tuning on a small amount of high-quality personalized alignment data.

\begin{table}[ht]\footnotesize\setlength{\tabcolsep}{4pt}
    \centering
    \caption{Pearson, Spearman, and Kendall correlations with human ratings for each $(\bm{x}, u)$ pair on \dataseta. We use three reviews ($k=3$) to represent reviewers' preferences. All results have a p-value less than 0.05. \method-7b is comparable to GPT-4 and \method-13b significantly outperforms GPT-4.}
    \begin{tabular}{lccc}
    \toprule
    & \textbf{Pearson} & \textbf{Spearman} & \textbf{Kendall} \\
    \midrule
    Reviewer Avg. & 0.301 & 0.302 & 0.230 \\
    Matrix Factorization & 0.308 & 0.313& 0.269 \\
    \midrule
    LLaMA-2-7b & 0.146 & 0.117 & 0.094 \\
    LLaMA-2-13b & 0.172 & 0.182 & 0.147 \\
    LLaMA-2-70b & 0.214 & 0.232 & 0.181 \\
    GPT-4 & 0.315 & 0.312 & 0.253 \\
    \midrule
    $\method$-7b & 0.307 & 0.329 & 0.263 \\
    $\method$-13b & \textbf{0.345} & \textbf{0.368} & \textbf{0.293} \\
    \bottomrule
    \end{tabular}%
  \label{tab:main_movie}%
\end{table}

\begin{table}[htbp]\footnotesize\setlength{\tabcolsep}{2pt}
  \centering
  \caption{The comparison of the generated explanation and the human-written review on \dataseta. A higher score indicates a better alignment between the generated explanation and the human reference. The reviews generated by \method are more similar to the human-written reviews. }
    \begin{tabular}{lcccc}
    \toprule
          & \textbf{BLEU}  & \textbf{ROUGE} & \textbf{BERTScore} & \textbf{BARTScore} \\
    \midrule
    LLaMA-7b & 2.213 & 0.253 & 0.829 & -9.049 \\
    LLaMA-13b & 2.847 & 0.262 & 0.833 & -9.228 \\
    LLaMA-70b & 3.014 & 0.256 & 0.832 & -8.538 \\
    GPT-4 & 3.040 & 0.252 & 0.831 & -6.853 \\
    \midrule
    \method-7b & 3.988 & 0.292 & \textbf{0.834} & -6.741 \\
    \method-13b & \textbf{4.108} & \textbf{0.294} & \textbf{0.834} & \textbf{-6.577} \\
    
    \bottomrule
    \end{tabular}%
  \label{tab:reviewtext}%
\end{table}%

\begin{table*}[htbp]\setlength{\tabcolsep}{4pt}\footnotesize
  \centering
  \caption{Fine-grained prediction accuracy for each $(\bm{x}, u, \bm{a})$ on \datasetb with $k=1$. \method-7b and \method-13b were trained on all aspects. \method outperforms all baselines in all aspects. The p-value for t-test are smaller than 0.05.}
    \begin{tabular}{lccccc|c}
    \toprule
          & \textbf{Interestingness} & \textbf{Adaptability} & \textbf{Surprise} & \textbf{Character} & \textbf{Ending} & \textbf{Average} \\
    \midrule
    Reviewer Avg. & 0.466 & 0.478 & 0.460 & 0.469 & 0.515 & 0.477 \\
    \midrule
    LLaMA-2-7b & 0.466 & 0.491 & 0.453 & 0.481 & 0.503 & 0.479 \\
    LLaMA-2-13b & 0.422 & 0.451 & 0.477 & 0.481 & 0.517 & 0.470 \\
    LLaMA-2-70b & 0.517 & 0.507 & 0.431 & 0.505 & 0.545 & 0.501  \\
    GPT-4 & 0.502 & 0.496 & 0.596 & 0.506 & 0.543 & 0.529  \\
    \midrule
    $\method$-7b & 0.572 & 0.565 & \textbf{0.619} & 0.565 & 0.560 & 0.576 \\
    $\method$-13b & \textbf{0.621} & \textbf{0.570} & 0.616 & \textbf{0.607} & \textbf{0.597} & \textbf{0.602} \\
    \bottomrule
    \end{tabular}%
  \label{tab:main_anno}%
\end{table*}%

\textbf{\method's Explanation of Scale Rating Aligns with Reviewer's Comments.}
We further investigate whether the explanations of the scalar ratings provided by \method align with the reviewer's comments. We use four widely used evaluation metrics in text generation to compare the explanation $\bm{e}_u$ with the ground-truth review text. These metrics include two lexical-similarity-based metrics: BLEU~\citep{papineni-etal-2002-bleu} and ROUGE~\citep{lin-2004-rouge}, and two model-based metrics: BERTScore~\citep{zhang2019bertscore} and BARTScore~\citep{NEURIPS2021_e4d2b6e6}\footnote{ROUGE-1 is used here. BARTScore is negative because it uses the average log-likelihood of the fine-tuned BART as the score.}. A higher score indicates that the generation is more similar to the reference. Table \ref{tab:reviewtext} shows that \method-7b and \method-13b outperform other baselines across all metrics. This suggests that \method can better model the preferences of a specific reviewer and generate a personalized review from this perspective.

\textbf{\method's Pairwise Rating Outperforms Other Baselines on All Aspects of \datasetb.} 
As shown in Table \ref{tab:main_anno}, \method achieves the best performance across all aspects. Compared to \method-13b, \method-7b achieves comparable performance on \texttt{Surprise} but falls behind on other aspects. Similarly, the vanilla LLaMA models struggle with modeling personal preferences and rarely surpass the simple baselines, achieving around 50\% accuracy on most aspects. This is partly due to having only $K=1$ historical review for each reviewer, which requires the LLMs to have a strong capability in inferring personal preferences from a limited profile. Meanwhile, although GPT-4 demonstrates relatively high accuracy in capturing \texttt{Surprise}, its performance in other aspects is not satisfactory.

\subsection{In-Depth Exploration}
We conduct additional experiments to investigate personalization alignment in \method. Some of these experiments are detailed in Appendix \ref{sec:a_analysis}.

\begin{figure}[ht]
    \centering
    \includegraphics[width=1\linewidth]{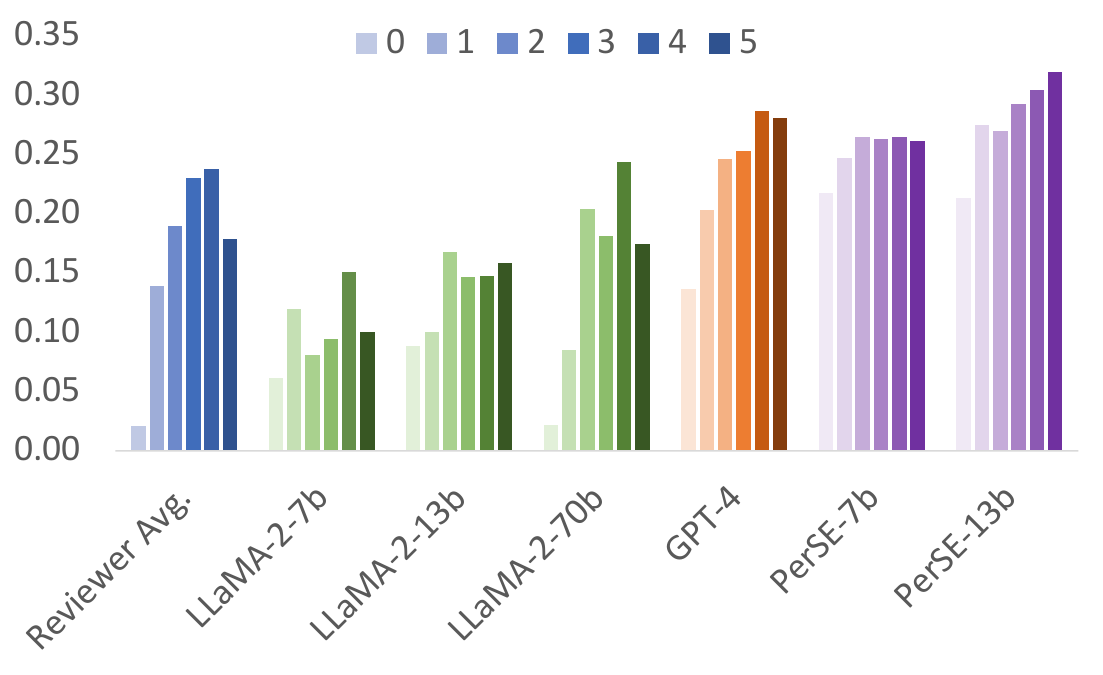}
    \caption{Kendall correlation on \dataseta with different numbers of historical reviews ($K$) in reviewer profile. Having more reviews benefits \method-13b, but the increased complexity may harm the performance of the vanilla LLaMA models.}
    \label{fig:review_number}
\end{figure}

\textbf{Personalized Evaluation of Open-ended Generation is Necessary.}
In Table \ref{fig:review_number}, $K=0$ indicates that there are no personalized examples in the instruction, meaning that $p_u$ is empty and only the query $\bm{x}$ is provided in the prompt. This represents a `one-score-fits-all' evaluation for open-ended generation, where the same score is predicted for all users given the same query. The poor performance of all baselines under this setting indicates that the variance in reviewers' preferences has a significant influence on the evaluation, highlighting that a `one-score-fits-all' approach is ineffective for evaluating open-ended generation. We further calculate the review score variance of \dataseta in Table \ref{tab:score_var} in the Appendix to emphasize the necessity of personalized alignment.

\textbf{With a Richer Reviewer Profile, \method Aligns Better with the Personal Preference.}
We further explore how many reviews are required to establish a reviewer's preference in Figure \ref{fig:review_number}. For \method-7b and \method-13b, we train the models on different subsets of \dataseta as shown in Table \ref{tab:dataset}. As observed, with an increase in the number of historical reviews $K$ in the profile, \method-13b's performance consistently improves. This indicates that after personalized alignment tuning, \method can better capture the personal preferences of historical reviewers. However, we also find that after 4 reviews, the performance of most baselines, including the average reviewer score baseline, declines. While a longer context provides more information about personal preferences, it also introduces challenges due to increased context complexity and noise. Therefore, we assume that increasing the number of historical reviews beyond a certain point may not further enhance \method's performance.

\begin{figure}[ht]
    \centering
    \includegraphics[width=1\linewidth]{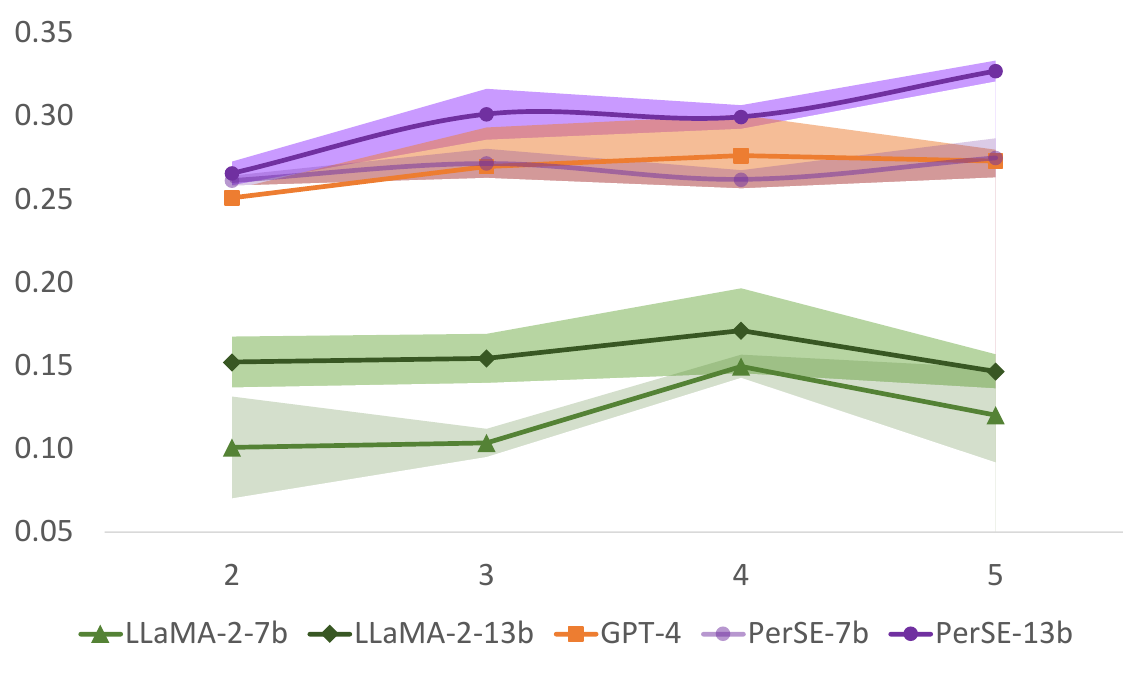}
    \caption{Kendall correlation on \dataseta with different orders of reviews. The shadow indicates the variance while the line is the average performance among three trials. \method is more stable than baselines.}
    \label{fig:review_order}
\end{figure}

\textbf{More Historical Reviews Make \method More Robust.}
\label{sec:k}
Previous studies have shown that LLMs are sensitive to the perturbation of in-context examples~\citep{lu-etal-2022-fantastically}. To investigate how robust \method is to the order of historical reviews, we randomly shuffle the historical reviews in the profile and rerun the experiments three times. We present the average performance with lines and the standard deviation as shaded regions in Figure \ref{fig:review_order}. We can see that \method-13b consistently outperforms other baselines on average and has a smaller shaded region, indicating that \method is more robust to changes in the order of the profile than the other baselines. 
Furthermore, as the number of reviews increases, \method's performance converges, suggesting that its performance is more stable with a larger profile. This implies that \method better captures the reviewer's preference and can coherently provide personalized scores for new queries without being affected by the order of reviews. In contrast, the vanilla LLaMA-2 models are more sensitive to order, showing a larger variance in the shaded regions.

\begin{figure}
    \centering
    \includegraphics[width=1\linewidth]{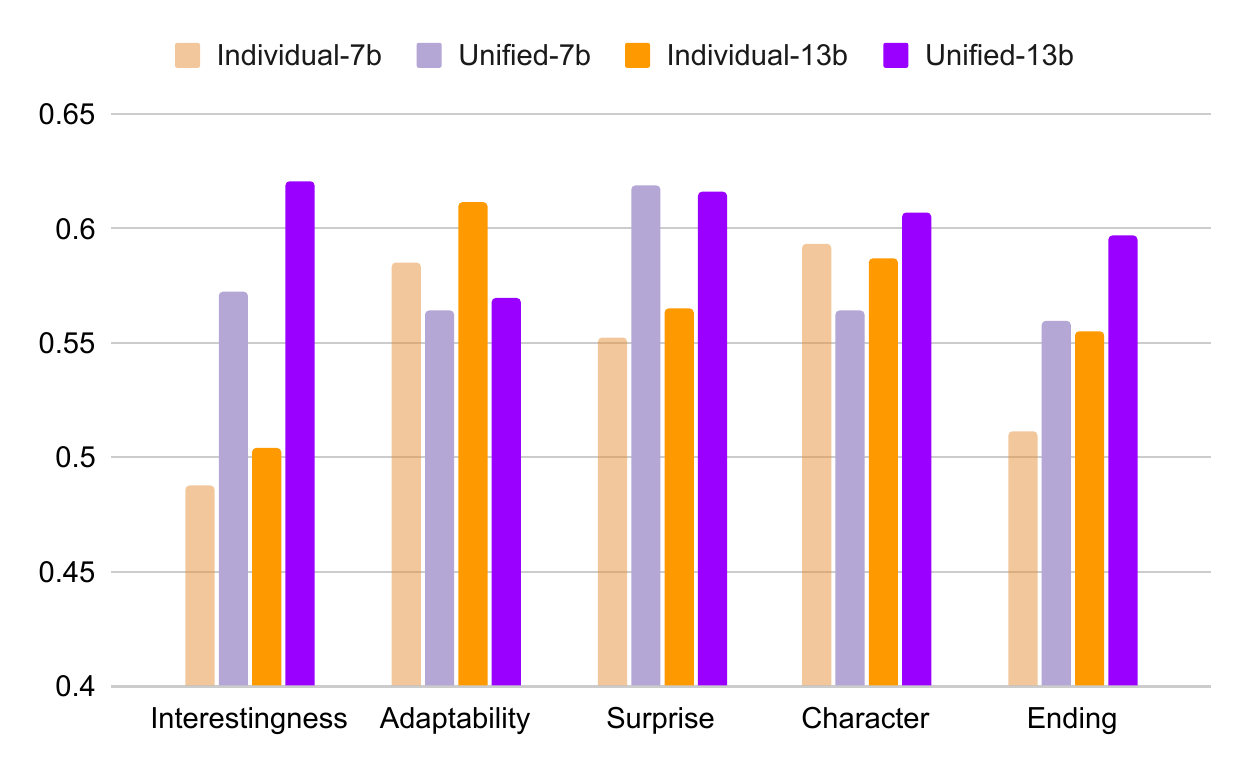}
    \caption{Accuracy of the unified and separate models on \datasetb. Unified training improves performance.}
    \label{fig:unified}
\end{figure}

\textbf{Training with Ratings from Different Aspects Helps \method Better Understand Preferences.}  
We investigate the influence of joint training on different aspects using \datasetb. We compare the performance of a unified model, trained on all aspects together, with that of models trained separately on each aspect. As illustrated in Figure \ref{fig:unified}, joint training improves performance in most aspects, as exposure to different aspects allows the models to benefit from one another. For example, the capabilities to capture \texttt{Interestingness} and \texttt{Surprise}, as well as to evaluate the quality of \texttt{Ending}, are weaker in the individual setting. However, these aspects are enhanced during joint training, resulting in significant improvement.
For separate models, they are better at capturing preferences for \texttt{Adaptability} and \texttt{Character Development}. We believe these two aspects are related to the plot's setting, which is more structured. This structure may lead to a clearer preference that is easier to capture with single-aspect data.

\textbf{\method Generalizes Well to Other Domains without Fine-tuning.}  
We evaluate the generalization of \method by applying it to a new domain (Amazon book review\footnote{https://nijianmo.github.io/amazon/index.html}) in a zero-shot manner. We recreate a personalized dataset from the Amazon book reviews for scalar rating using the pipeline described in Section \ref{sec:curate}. Detailed data statistics can be found in Appendix \ref{sec:a_data_collection}. To fit the scoring range of the Amazon dataset, which is 1 to 5, we calibrate \method's predictions (originally 1 to 10). We directly use \method, tuned on \dataseta, to predict personalized reviews and scores for each book based on the user's preference. As shown in Table \ref{tab:amazon}, \method outperforms other baselines even without fine-tuning on the new domain, indicating that \method can be effectively applied to new domains with limited or no fine-tuning data.

\textbf{Personalized Tuning in \method Works Well with Other LLMs.}  
In the main experiment, we use LLaMA-2 as the backbone LLM. Here, we also investigate whether our training process is applicable to other LLMs. We use the same data and training method to fine-tune Mistral 7B~\citep{jiang2023mistral}. Results in Table \ref{tab:mistral} show that our method enhances the capability of the Mistral 7B model in both in-domain and out-of-domain settings.

\begin{table}[htbp]\footnotesize\setlength{\tabcolsep}{4pt}
  \centering
  \caption{Zero-shot performance on Amazon book review. The experimental setting is the same as Table \ref{tab:main_movie}.}
    \begin{tabular}{lccc}
    \toprule
       & \textbf{Pearson} & \textbf{Spearman} & \textbf{Kendall} \\
    \midrule
    Reviewer Avg. & 0.146 & 0.180 & 0.177 \\
    \midrule
    LLaMA-7b & 0.066 & 0.127 & 0.124 \\
    LLaMA-13b & 0.070 & 0.122 & 0.112 \\
    LLaMA-70b & 0.116 & 0.150 & 0.146 \\
    GPT-4 & 0.152 & 0.165 & 0.162 \\
    \midrule
    $\method$-7b & 0.170 & 0.238 & 0.219 \\
    $\method$-13b & \textbf{0.217} & \textbf{0.247} & \textbf{0.237} \\
    \bottomrule
    \end{tabular}%
  \label{tab:amazon}%
\end{table}%

\begin{table}[ht]\footnotesize\setlength{\tabcolsep}{4pt}
    \centering
    \caption{Mistral-7b-based \method outperforms the original pre-trained model and achieves comparable performance with \method-7b. The setting is the same as Table \ref{tab:main_movie}.}
    \begin{tabular}{lccc}
    \toprule
    & \textbf{Pearson} & \textbf{Spearman} & \textbf{Kendall} \\
    \midrule
     \multicolumn{4}{c}{In-domain: \dataseta} \\
    \midrule
    Mistral 7B & 0.166 & 0.128 & 0.106 \\
    $\method$-Mistral & 0.302 & 0.320 & 0.250 \\
    \midrule
     \multicolumn{4}{c}{Out-of-domain: Amazon Book Review} \\
    \midrule
    Mistral 7B & 0.088 & 0.102 & 0.098 \\
    $\method$-Mistral & 0.170 & 0.218 & 0.204 \\
    \bottomrule
    \end{tabular}%
  \label{tab:mistral}%
\end{table}

\begin{figure*}[t]
    \centering
    \includegraphics[width=1\linewidth]{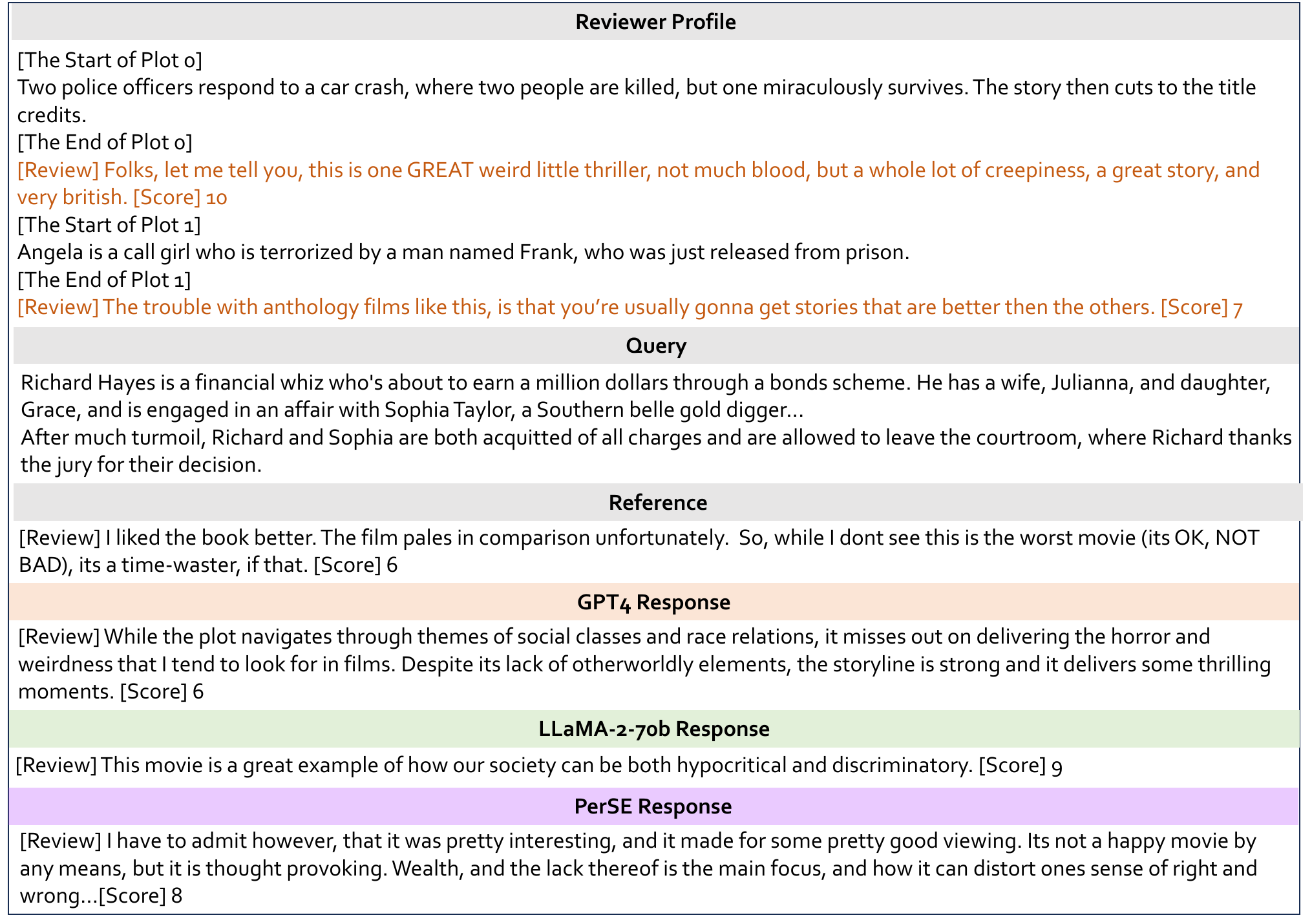}
    \caption{An example for evaluating the individual story from the given reviewer's preference. The reference is the ground-truth reviews given by this reviewer. More cases are shown in the Appendix \ref{sec:a_analysis}.}
    \label{fig:case_study}
\end{figure*}

\textbf{Compared with \method's Personalized Alignment, GPT-4's Generic Assessment Fails to Align with Specific Reviewer Preferences.} In Figure \ref{fig:case_study}, we present an example from \dataseta. The annotated reviews reveal that this reviewer is critical of plots and particularly values novelty. However, even with this reviewer's preferences provided, GPT-4 predicts a positive review. We believe this is because GPT-4 is overly aligned towards safety and harmlessness, making it cautious in giving negative responses. While LLaMA-2-70b is stricter and assigns a score of 4, \method focuses more on the consistent terribleness and assigns a score of 3, which aligns more closely with the reviewer's true score. Moreover, we find that, unlike most people, this reviewer does not prioritize complicated themes. Despite this, GPT-4's "one-size-fits-all" evaluation offers a high score for such themes. \method, on the other hand, pays more attention to this reviewer's visual preferences, providing a more reviewer-specific rating. This indicates that \method can better evaluate stories based on personalized preferences rather than relying on a general and universally applicable evaluation principle without any personalized preference.

\section{Conclusion and Discussion}
\label{sec:conclusion}
In this paper, we focus on learning personalized alignment in evaluating open-ended generation. We introduce \method, an LLM-based evaluation model that provides an interpretable evaluation from the perspective of an unseen reviewer. It infers the reviewer's preferences based on their profile and applies these preferences to the evaluation of open-ended generation. In addition to providing a score, \method also offers a detailed explanation. By being instruction-tuned on personalized alignment data, the LLaMA-2-based \method outperforms GPT-4 in both scalar and pairwise ratings. Our comprehensive analysis of personalized alignment underscores the importance of personalized fine-tuning to avoid over-alignment with common human values, often a result of RLHF. The interpretability of \method makes it particularly well-suited for personalized generation and recommendation systems.

\section*{Limitation}
\label{sec:limitation}
While this research makes notable strides in addressing the challenge of personalized evaluation, it is not without limitations. For instance, we assume that preferences remain consistent across prior reviews, which may not account for changes in preferences in real-world scenarios. It would be interesting to model how preferences shift over time and evaluate content based on potential future preferences. Additionally, 
we demonstrate that with instruction tuning on personalization data, the smaller LLaMA-2 can outperform the larger GPT-4. More exploration could be conducted with large-scale LLMs to assess the scalability of our method.

\section*{Ethics Statement}
\label{sec:ethics}
As we conduct extensive research to enhance and personalize the capabilities of Large Language Models (LLMs) such as \method, we remain ever-conscious of the ethical implications of our work.

One ethical concern is ensuring fairness and avoiding potential bias in the personalization of LLMs. While \method aims to evaluate content based on individual preferences, we carefully construct the instruction data to mitigate potential undesirable behaviors during fine-tuning. We also enhance the transparency of personalized evaluations by introducing interpretable metrics, as suggested by \citet{kirk2023personalisation}.

Another ethical consideration relates to privacy and consent. The two datasets, \dataseta and \datasetb, are reproduced from existing publicly released datasets MPST~\citep{KAR18.332,kar-etal-2020-multi} and DOC~\citep{yang-etal-2023-doc}, under their respective licenses. They are sourced ethically, and we always respect individual privacy. All data used is aggregated and anonymized to safeguard personal information.

We remain committed to conducting our research responsibly, adhering to ethical guidelines, to ensure that our contributions to AI advancements promote transparency, fairness, and respect for privacy.

\bibliography{custom}

\newpage
\appendix

\label{sec:appendix}

\section{Contamination Issues in Evaluation}
\label{sec:a_mem}

\begin{table*}[ht]\small
    \centering
    \caption{Performance of GPT-4 in predicting a scalar rating for a single synopsis. Percent is the percentage of each type of synopsis (raw/anonymized/summarized) being recognized as `memorized’, or `unmemorized’. Memorization heavily affects performance, but its impact decreases with anonymization and summarization.}
        \begin{tabular}{clcccc}
        \toprule
          &       & \multicolumn{1}{c}{\textbf{Pearson}} & \multicolumn{1}{c}{\textbf{Spearman}} & \multicolumn{1}{c}{\textbf{Kendall}} & \multicolumn{1}{c}{\textbf{Percent}} \\
        \midrule
        \multirow{3}[2]{*}{\textbf{Memorized}} & Raw & 0.680 & 0.718 & 0.590 & 84.5\% \\
              & Anonymized & 0.682 & 0.680 & 0.548 & 57.5\% \\
              & Summarized  & 0.621 & 0.648 & 0.552 & 27.0\% \\
        \midrule
        \multirow{3}[2]{*}{\textbf{Un-memorized}} & Raw & 0.460 & 0.470 & 0.364 & 15.5\% \\
              & Anonymized & 0.216 & 0.289 & 0.222 & 42.5\% \\
              & Summarized  & 0.232 & 0.271 & 0.217 & 72.5\% \\
        \bottomrule
        \end{tabular}%
      \label{tab:single_bias}%
\end{table*}

Many existing evaluation datasets have been exposed to LLMs during their pretraining, leading to contamination issues when assessing model performance on these datasets. LLMs might achieve excellent performance on contaminated cases by memorizing the ground truth but perform poorly on others, rendering the assessment unreliable.

To address this, we investigate how such contamination affects LLMs when evaluating open-ended generation under three evaluation settings: scalar rating, pairwise rating, and personalized alignment. We evaluate GPT-4's performance on the IMDb dataset\footnote{\url{https://developer.imdb.com/non-commercial-datasets/}}. This movie dataset includes the synopsis and a score (1 to 10) for each movie. GPT-4 is tasked with identifying the movie based on the synopsis and evaluating the quality of the synopsis. We consider a movie to be memorized by GPT-4 if it can correctly predict the movie title from the synopsis. Additionally, we evaluate GPT-4's performance on a preprocessed version of the IMDb dataset that incorporates anonymization and summarization as described in Section \ref{sec:curate}.

\subsection{Memorization in Scalar Rating}

We ask GPT-4 to predict the score for a given synopsis. We then group the results based on the memorization status into two sets: 'Memorized' and 'Un-memorized'. We calculate the correlation between the predicted scores and the ground-truth scores. The results are presented in Table \ref{tab:single_bias}.

GPT-4's predictions show a very high correlation with the ground-truth scores for memorized cases. However, the performance drops dramatically for un-memorized cases. This indicates that the memorization issue makes the evaluation of GPT-4's performance unreliable: rather than analyzing the quality of the movie based on the given synopsis, GPT-4 relies on its memory of the score. We also find that the percentage of memorized cases significantly decreases after applying anonymization and summarization, demonstrating their effectiveness in alleviating memorization issues.

\begin{table*}[htbp]\footnotesize\setlength{\tabcolsep}{4pt}
   \centering
    \caption{GPT-4 in comparing two synopsis. Cons. is the percentage of consistent results when swapping the order. Bias First is the percentage where GPT-4 favors the first answer more than the ground truth. Overall, memorization leads to greater position bias and lower consistency. }
    \begin{tabular}{clcccc}
    \toprule
          &       & \multicolumn{1}{c}{\textbf{Accu. $\uparrow$} } & \multicolumn{1}{c}{\textbf{Cons. $\uparrow$}} & \multicolumn{1}{c}{\textbf{Bias First $\downarrow$}} & \multicolumn{1}{c}{\textbf{Percent}} \\
    \midrule
    \multirow{3}[2]{*}{\makecell{\textbf{Both} \\ \textbf{Memorized}}} & Raw & 0.714 & 63.0\% & 16.5\% & 91.0\% \\
          & Anonymized & 0.712 & 60.7\% & 17.8\% & 73.0\% \\
          & Summarized  & 0.753 & 73.4\% & 12.9\% & 42.5\% \\
    \midrule
    \multirow{3}[2]{*}{\makecell{\textbf{One} \\ \textbf{Memorized}}} & Raw & 0.778 & 78.9\% & -11.1\% & 9.0\% \\
          & Anonymized & 0.804 & 71.7\% & -6.5\% & 23.0\% \\
          & Summarized  & 0.632 & 82.4\% & 1.5\% & 34.0\% \\
    \midrule
    \multirow{3}[2]{*}{\makecell{\textbf{Neither} \\ \textbf{Memorized}}} & Raw & /     & /     & /     & 0.0\% \\
          & Anonymized & 0.500 & 62.5\% & 25.0\% & 4.0\% \\
          & Summarized  & 0.660 & 85.1\% & 4.3\% & 23.5\% \\
    \bottomrule
    \end{tabular}%
    \label{tab:pair_bias}%
\end{table*}

\subsection{Memorization in Pairwise Rating} 

We create 200 movie pairs, each consisting of two movie synopses with scores differing by at least 1 point. We ask GPT-4 to identify the titles and predict which synopsis is better. 
We calculate prediction accuracy (Accu.), consistency (Cons.), and bias towards the first synopsis. Consistency measures how many judgments remain consistent after changing the order of the two plots. Bias towards the first is defined as an inappropriate preference for the first synopsis. It is calculated by subtracting the percentage where GPT-4 favors the first plot from the true percentage of the first being better.

Results are reported in Table \ref{tab:pair_bias}. Similarly to scalar ratings, the neither-memorized group exhibits much lower accuracy compared to the other two groups, despite maintaining the main plot points. This indicates that memorization can lead to misleadingly high performance in evaluation. 
When GPT-4 memorizes one of the two plots, it is more consistent in its judgment and shows a lower position bias. This occurs because GPT-4 favors the memorized plot regardless of its order in the pair. The use of anonymization and summarization reduces the both-memorized cases to 42.5\% and increases the neither-memorized cases to 23.5\%.

We further calculate the 'Bias memorized' by subtracting the percentage that GPT-4 favors the memorized plot from the true percentage where this plot is actually better. In Table \ref{tab:bias_known}, we observe that for all raw, anonymized, and summarized plots, GPT-4 shows a clear tendency to choose the memorized plot. This tendency is more pronounced in the summarized plots. We believe this is because data processing increases the uncertainty of the prediction, causing the model to be more conservative and rely on what it has memorized. However, GPT-4 also demonstrates high consistency and low position bias in the 'neither memorized' group (see Table \ref{tab:pair_bias}), indicating that when evaluating two novel stories, it can overcome the effects of memorization and assess based on the actual plots.

\begin{table}[ht]\small
    \centering
    \caption{Prediction on `One Memorized' Group in pairwise comparison of GPT-4. The `Raw', `Anonymized', and `Summarized' have the same meaning in Table \ref{tab:pair_bias}. }
    \begin{tabular}{lc}
    \toprule
          & \multicolumn{1}{l}{Bias Memorized} \\
    \midrule
    Raw & 0.222  \\
    Anonymized & 0.283 \\
    Summarized  & 0.397 \\
    \bottomrule
    \end{tabular}%
  \label{tab:bias_known}%
\end{table}

\subsection{Memorization in Personalization}
\begin{figure}
    \centering
    \includegraphics[width=1\linewidth]{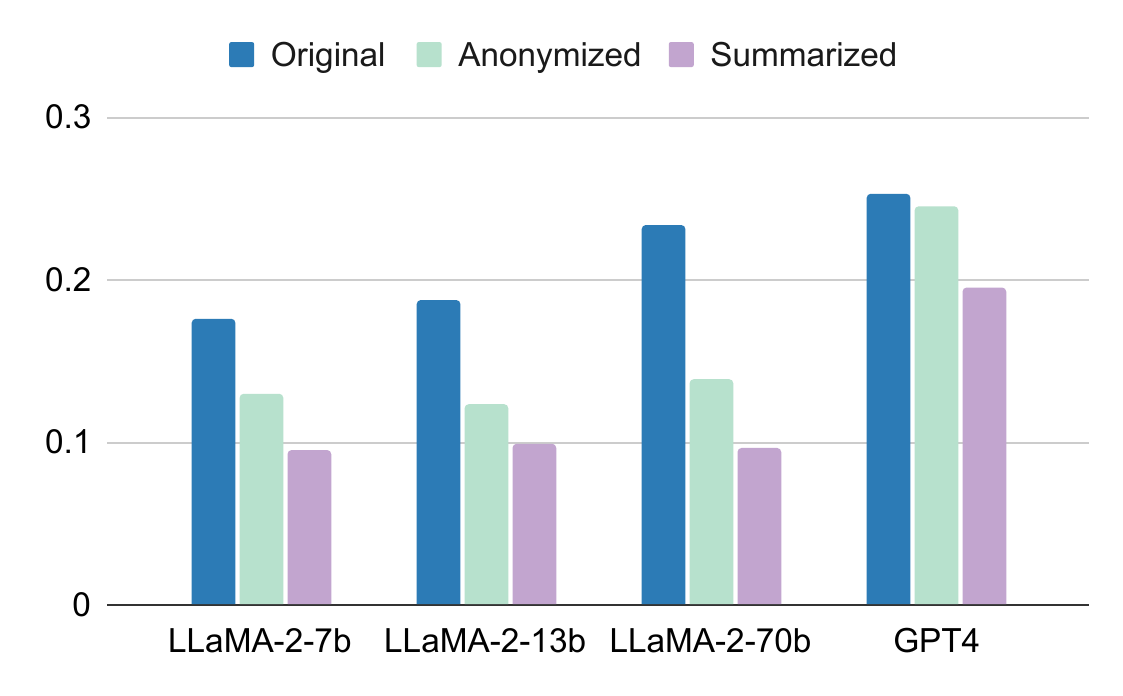}
    \captionof{figure}{Kendall correlation between the LLM's personalized prediction with human ratings. Personalized predictions of all LLMs are also affected by memorization.}
    \label{fig:contamination}
\end{figure}

We also explored the influence of memorization in personalized alignment during evaluation. We evaluate the performance of GPT-4 and vanilla LLaMA-2, as described in Section \ref{sec:eval}, on \dataseta. We use $K=1$ and calculate the Kendall correlation between human ratings and the predicted scores, as shown in Figure \ref{fig:contamination}.

Similarly, LLMs achieved a high correlation with human ratings on the original synopses, but their performance degraded after anonymization and summarization. Although the main plots remain the same, with only slight differences in recognizable details, this greatly affected the results. Both experiments highlight that memorization introduces significant bias in LLM-based evaluation models, rendering them unreliable for both general and personalized evaluations.

Overall, for LLM-based evaluation, contamination results in an unfairly high rating for exposed plots compared to unexposed ones.

\section{Constructing Personalized Alignment Datasets}
\label{sec:a_data_collection}

\begin{figure*}[ht]
    \centering
    \includegraphics[width=1\linewidth]{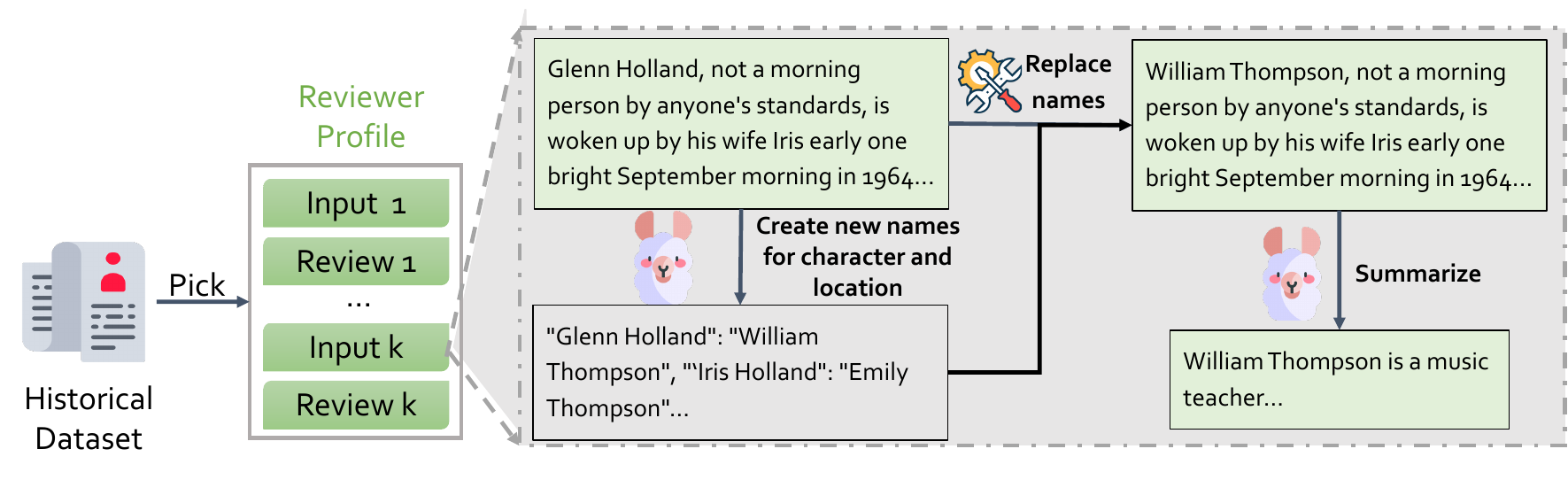}
    \captionof{figure}{The flowchart to construct our dataset. We use oasst-30b~\citep{kopf2023openassistant}, an instruction-tuned LLaMA-based model for anonymization and summarization. The prompts are listed in Figure \ref{fig:prompts}.}
    \label{fig:flowchart}
\end{figure*}

\begin{figure*}[ht]
    \centering
    \includegraphics[width=1\linewidth]{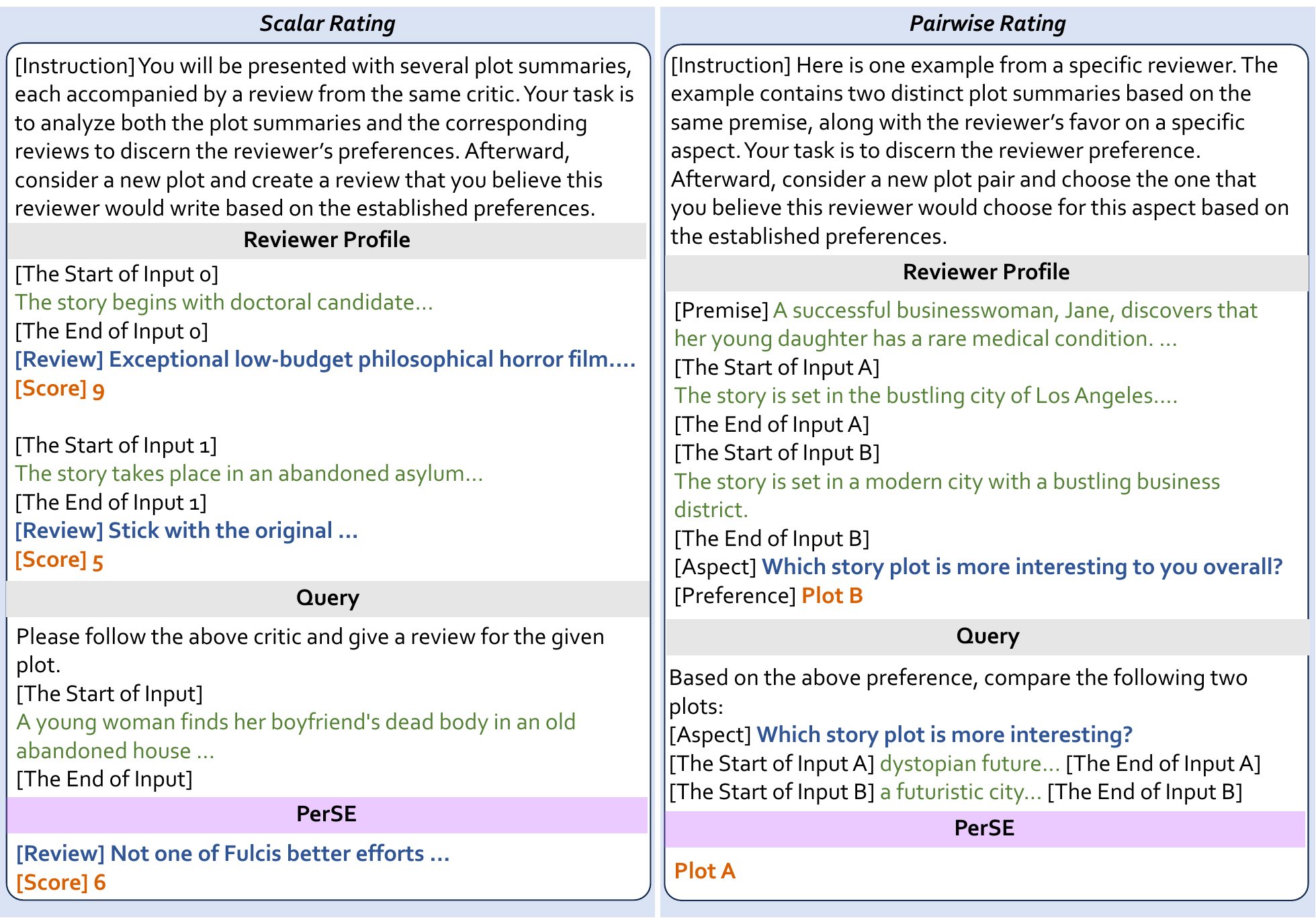}
    \captionof{figure}{The demonstrate of \method. The input is in green, the detailed review and fine-grained aspects are in blue, and the review scores are in orange. }
    \label{fig:perse}
\end{figure*}

To alleviate contamination issues, we create a less biased personalized evaluation dataset by anonymizing famous characters and summarizing existing plots. Our pipeline is illustrated in Figure \ref{fig:flowchart}. We use oasst-30b~\citep{kopf2023openassistant}, a 30B LLaMA-based model fine-tuned on OpenAssistant Conversations for alignment, to anonymize and summarize the plots.

Specifically, we anonymize the raw plot by asking the LLMs to identify character and local names and then create new names for them. Based on the JSON mapping generated, we replace those names with new ones. We do not directly ask LLMs to replace names to avoid potential hallucinations during the replacement. For characters with the same family names, LLMs can create new character names that still share the same last names (though not the original last names). For example, 'Glenn Holland' and 'Iris Holland' are mapped to 'William Thompson' and 'Emily Thompson'.

\begin{figure*}[ht]
    \centering
    \begin{subfigure}[t]{0.32\textwidth}
        \includegraphics[width=\textwidth]{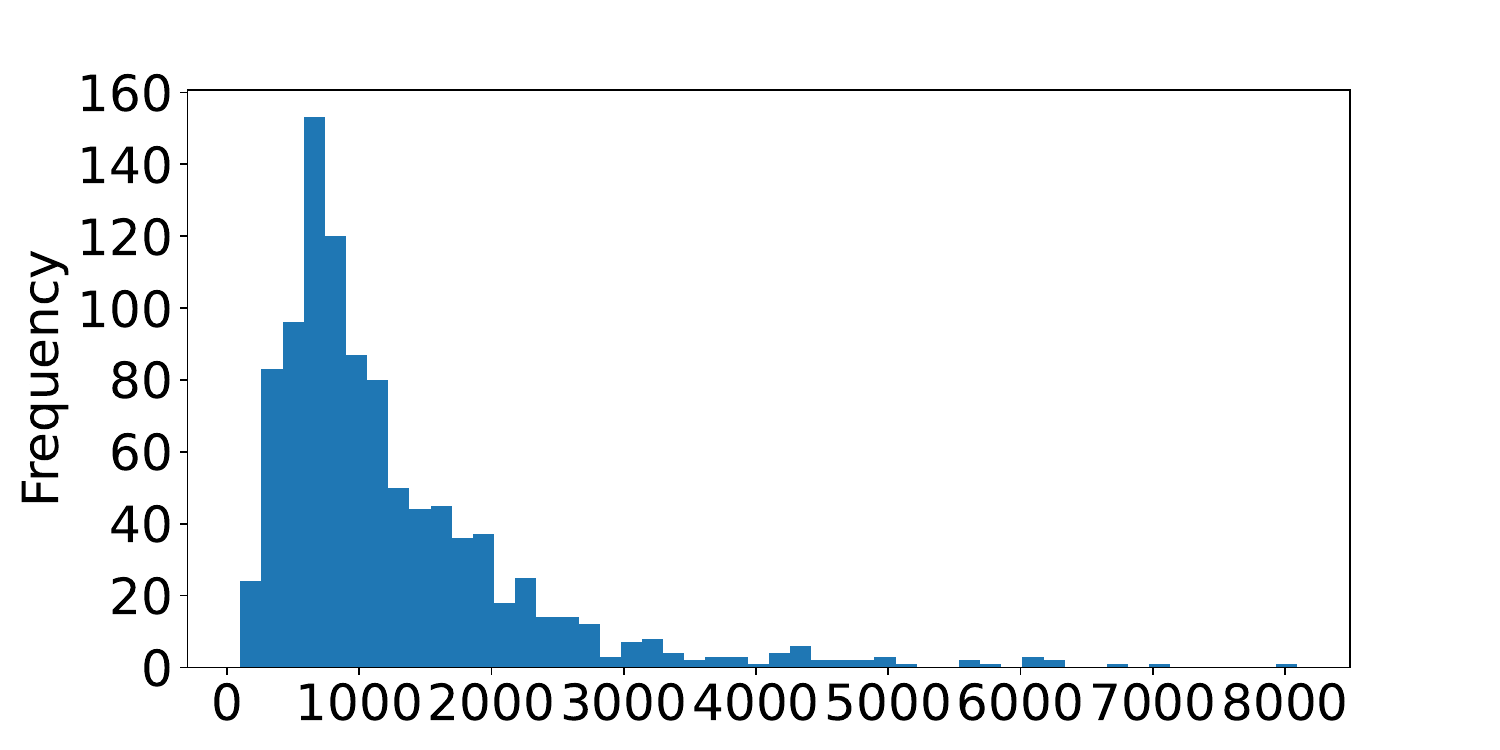}
        \caption{Raw movie length in MPST v2.}
        \label{fig:raw_movie}
    \end{subfigure}
    \hfill
    \begin{subfigure}[t]{0.32\textwidth}
        \includegraphics[width=\textwidth]{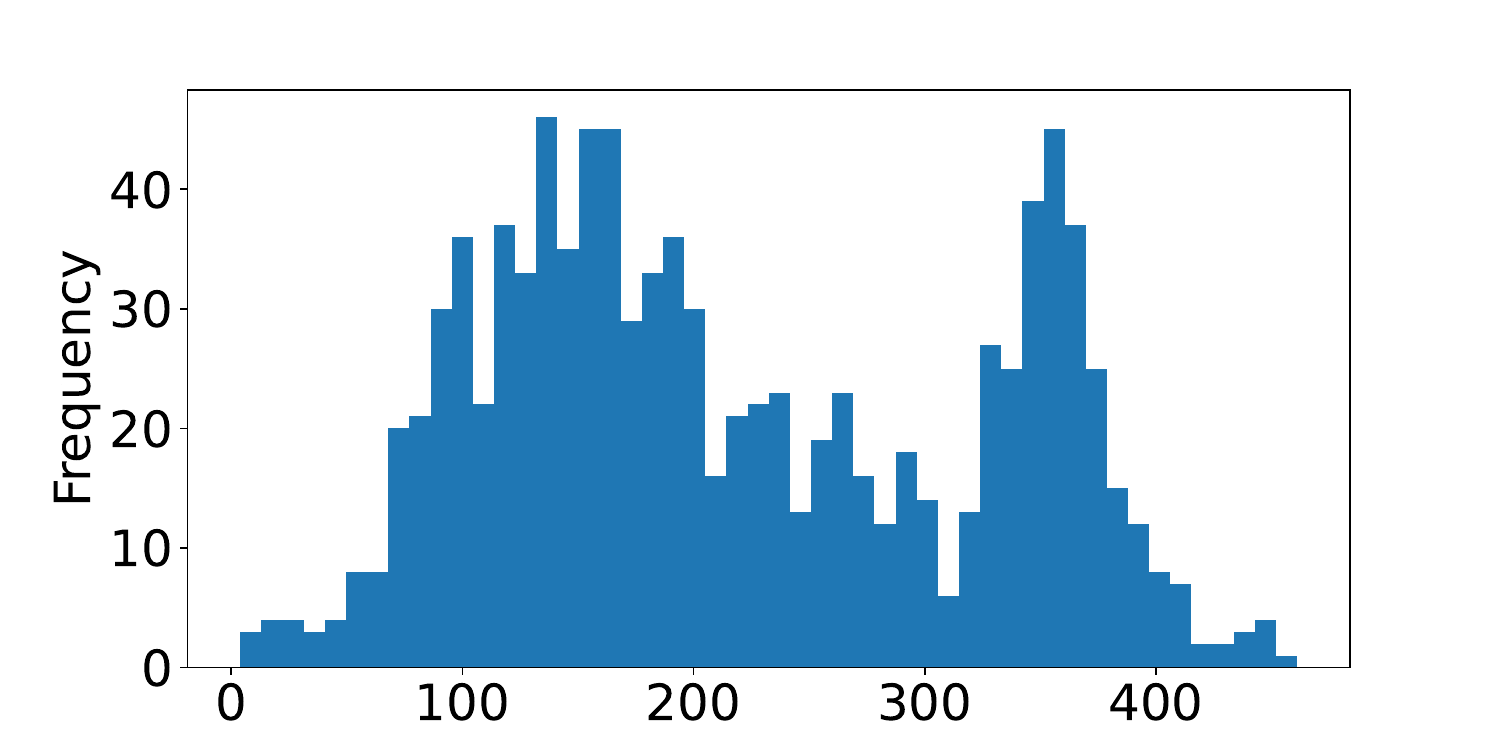}
        \caption{Movie length in \dataseta.}
        \label{fig:summ_movie}
    \end{subfigure}
    \hfill
    \begin{subfigure}[t]{0.32\textwidth}
        \includegraphics[width=\textwidth]{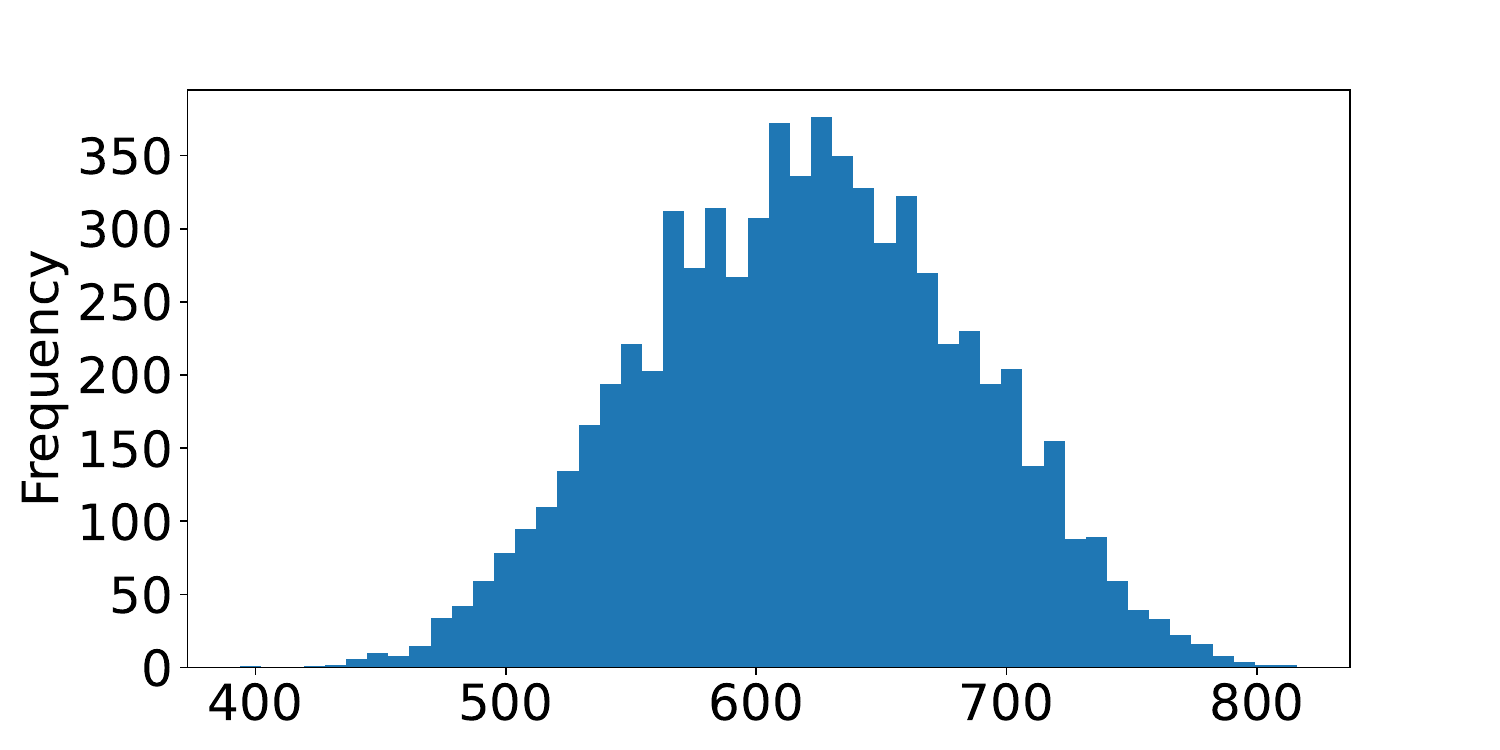}
        \caption{Story length in \datasetb.}
        \label{fig:raw_plot}
    \end{subfigure}
    \caption{Length Distribution of \dataseta and \datasetb. The x-axis is the length and the y-axis is the frequency.}
    \label{fig:length}
\end{figure*}

For \datasetb, we define five aspects based on the questions in \citet{yang-etal-2023-doc}:
\begin{enumerate}
[noitemsep, leftmargin=2em, topsep=1pt]
\item \texttt{Interestingness}: Which story plot is more interesting to you?  
\item \texttt{Adaptability}: In your opinion, which one of the plots above could generate a more interesting book or movie (when a full story is written based on it)?
\item \texttt{Surprise}: Which story plot created more suspense and surprise? 
\item \texttt{Character Development}: Which story's characters or events do you identify with or care for more? 
\item \texttt{Ending}: Which story has a better ending? 
\end{enumerate}
These aspects evaluate the three key elements in the story: Interestingness and Surprise for the plot, Character development for the character, and Ending and Adaptability for the setting. For each question, there are four options: plot A, and plot B, both are good, and neither is good. We remove the examples with the answer of `Both' and `Neither' because they do not show preference.

We illustrate the length distribution of the synopsis in \dataseta and the story in \datasetb in Figure \ref{fig:summ_movie} and \ref{fig:raw_plot}. For \dataseta, we also provide the length distribution of the raw plots in Figure \ref{fig:raw_movie}.

\begin{figure*}[h]
    \centering
    \includegraphics[width=1\linewidth]{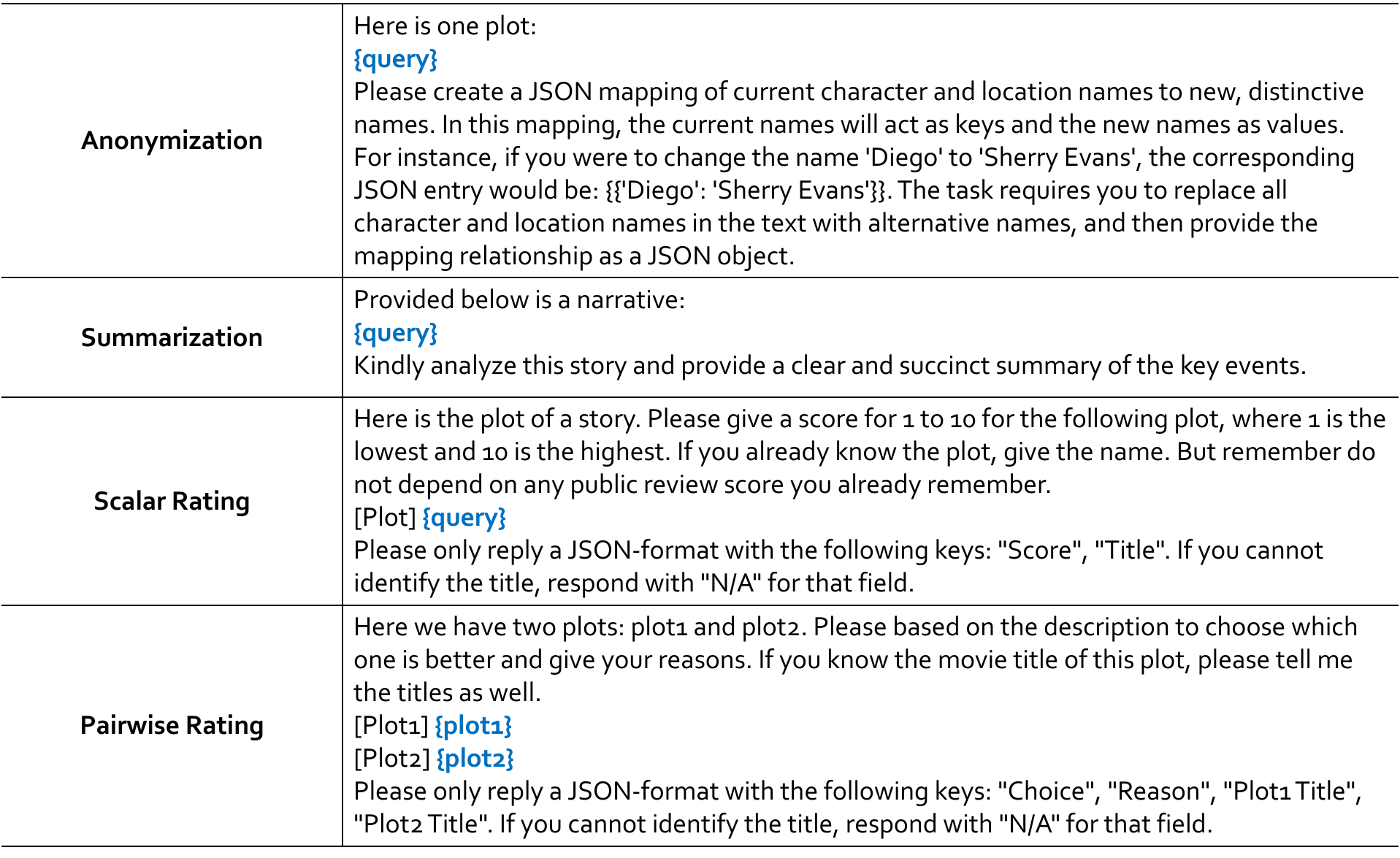}
    \caption{Prompts used in the data processing and investigation of contamination. The blue text is the placeholder for plots.}
    \label{fig:prompts}
\end{figure*}

\subsection{Variation of Score Preference across Reviewers}
We present the variation of score preferences in \dataseta in Table \ref{tab:score_var}. We computed the count of reviewers for each query, along with the average (mean) and standard deviation (std) of the reviewers' scores. The table shows that while the average scores for queries are almost identical, there is a considerable standard deviation, highlighting the differences in reviewer opinions. This underscores the need for an evaluation method that accounts for varying preferences.

\begin{table}[htbp]
  \centering
  \caption{Score variance between reviewers in \dataseta}
    \begin{tabular}{lccc}
    \toprule
       & \multicolumn{1}{c}{\# Review} & \multicolumn{1}{c}{Score Mean } & \multicolumn{1}{c}{Score STD} \\
    \midrule
    Train & 28.37 & 6.69 & 1.97 \\
    Test & 4.64 & 6.84 & 1.39 \\
    \bottomrule
    \end{tabular}%
  \label{tab:score_var}%
\end{table}%

\textbf{Amazon Book dataset}
Similar to \dataseta, we preprocess the Amazon book dataset to create a personalized version. We use the 5-core subset of the book domain, where every user and item has a minimum of 5 reviews. The original instances in the Amazon book dataset only include the book title without the content. Therefore, we use LLM-based retrieval to add a brief book description for each instance. Each example features one annotated review serving as the user profile ($K=1$). Ultimately, we create a personalized version of the Amazon book dataset, consisting of 120 evaluation examples.

\section{Prompts}
\label{sec:a_prompt}
We demonstrate the framework and prompts of \method in Figure \ref{fig:perse}. The prompts used in Appendix \ref{sec:a_mem} for addressing the contamination issue, as well as the prompts for anonymization and summarization, are listed in Table \ref{fig:prompts}.

\section{Training Details}
\label{sec:training}
Each model in our experiments was trained on 8 x 80G A100 GPUs with a learning rate of 1e-5. We set the batch size to 4 for \method-7b and 2 for \method-13b. The models converged after 2k and 6k steps on \dataseta, respectively.
We trained two unified models on \datasetb for all aspects by fine-tuning 7b and 13b LLaMA-2-chat. The models after 1k and 2k steps. It takes about 10 hours in total for these two models.
For the ablation study, we also trained one model for each aspect on \datasetb. Each model converged after 500 steps for 7b and 2k steps for 13b. The total training time was approximately 5 x 5 hours.

\section{More Case Studies}
\label{sec:a_analysis}

\begin{figure*}
    \centering
    \includegraphics[width=1\linewidth]{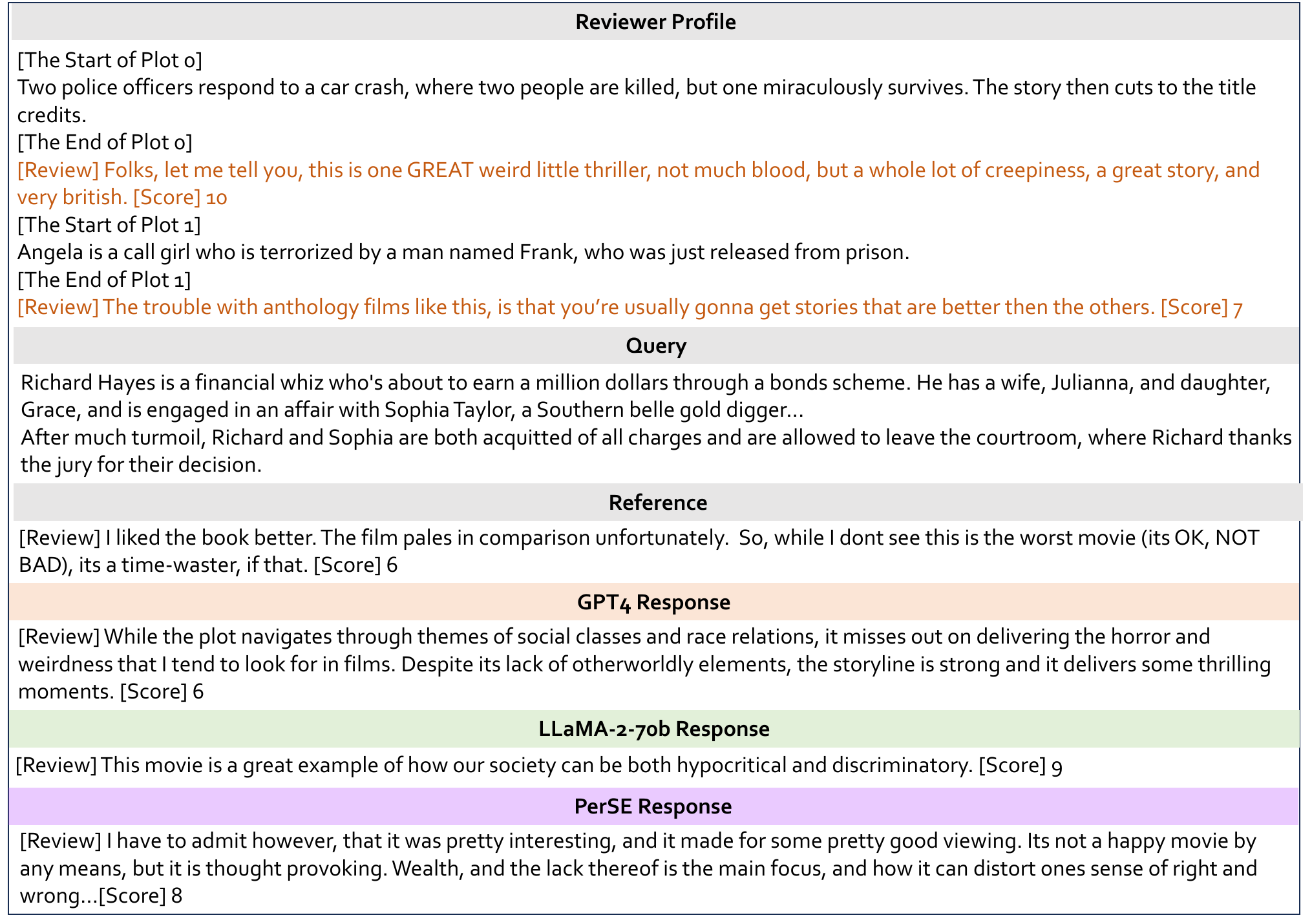}
    \caption{The score given by the reviewers on the new plot is very different from the comments with annotations. While LLaMA-2-70b and GPT-4 give a more similar score, \method is able to infer the preference and provide a score that is closer to the true score but far away from the annotated scores.}
    \label{fig:case_study_2}
\end{figure*}

\textbf{\method Infers Preferences Rather Than Copying Scores from Context.} In Figure \ref{fig:case_study_2}, we present another example from \dataseta. From the reviews, we can see that the reviewer enjoys horror elements. However, the new plot and its level of terror are unsatisfactory, leading the reviewer to give it a low score. Both GPT-4 and LLaMA-2-70b emphasize the horror theme and predict a high score for this plot. We suspect they are influenced by the high review scores in the reviewer's preference, overlooking the analysis of the new plot. In contrast, \method focuses on the dull aspects of the plot, aligning more closely with what the reviewer is concerned about. It assigns a score of 5, which differs from the existing review scores but is closer to the actual score the reviewer gave this plot.

\begin{figure*}
    \centering
    \includegraphics[width=1\linewidth]{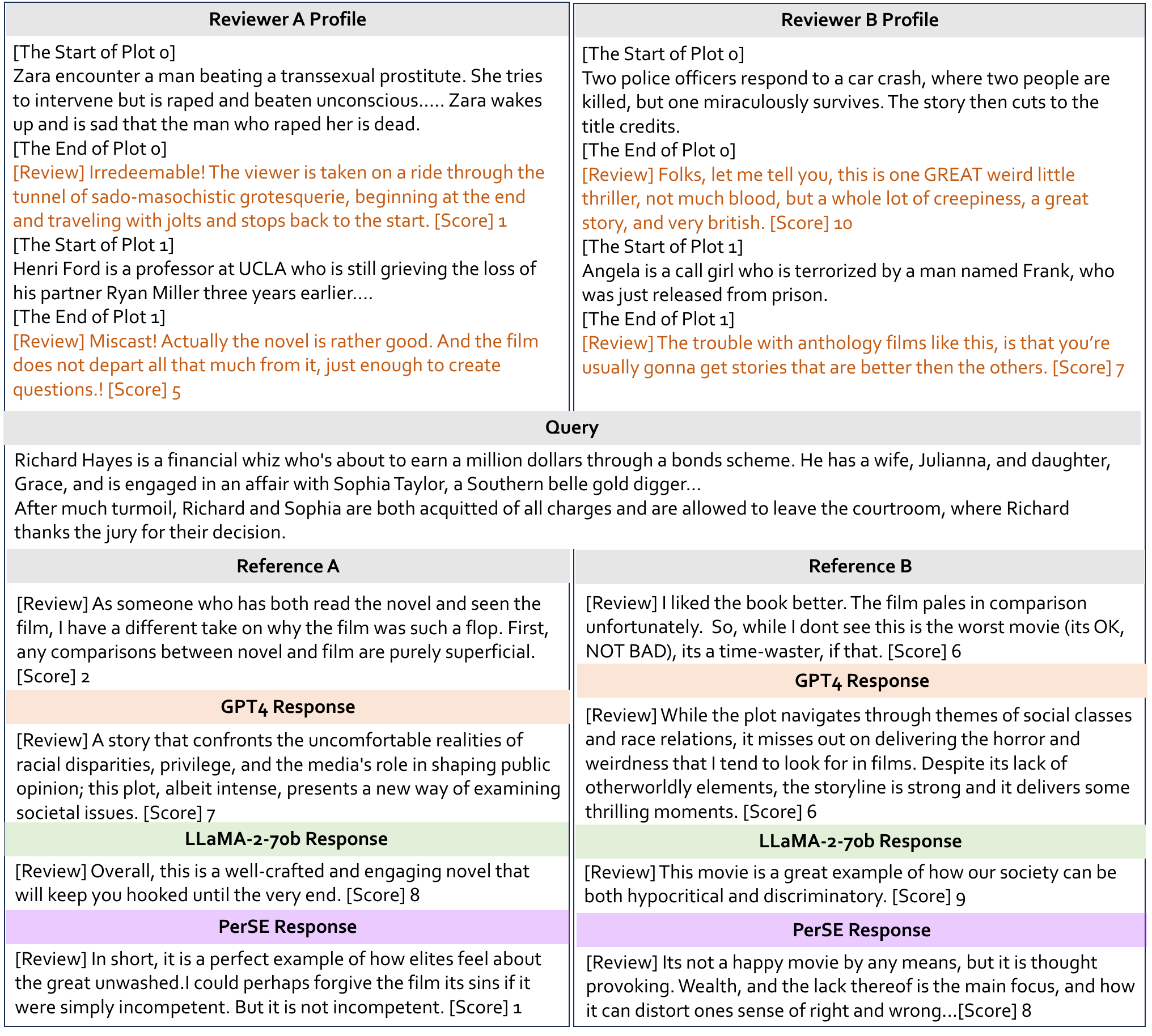}
    \caption{Reviews from two reviewers on the same plot. \method is able to give personalized scores based on preference. }
    \label{fig:case_study_34}
\end{figure*}

\textbf{\method Provides Diverse Reviews for the Same Plot Based on Different Preferences.} In Figure \ref{fig:case_study_34}, we illustrate the reviews of the same plot from two reviewers, A and B, each with different preferences. Both reviewers have read the book. Reviewer A is critical and has a high standard for good movies, leading to low scores in the annotated reviews. Consequently, he gives a score of 2 due to his disappointment with the movie adaptation.
In contrast, Reviewer B is relatively tolerant and tends to give high scores. Although the movie is much worse than the book, he still gives a score of 6. However, GPT-4 and LLaMA-2-70b assign similarly high scores in both cases, disregarding the reviewers' preferences. However, \method is capable of providing personalized scores for different reviewers, predicting a score of 1 for Reviewer A and 8 for Reviewer B. Although the predicted score for Reviewer B is not as close as GPT-4's, it reflects the positive attitude captured by \method.

\begin{figure*}
    \centering
    \includegraphics[width=1\linewidth]{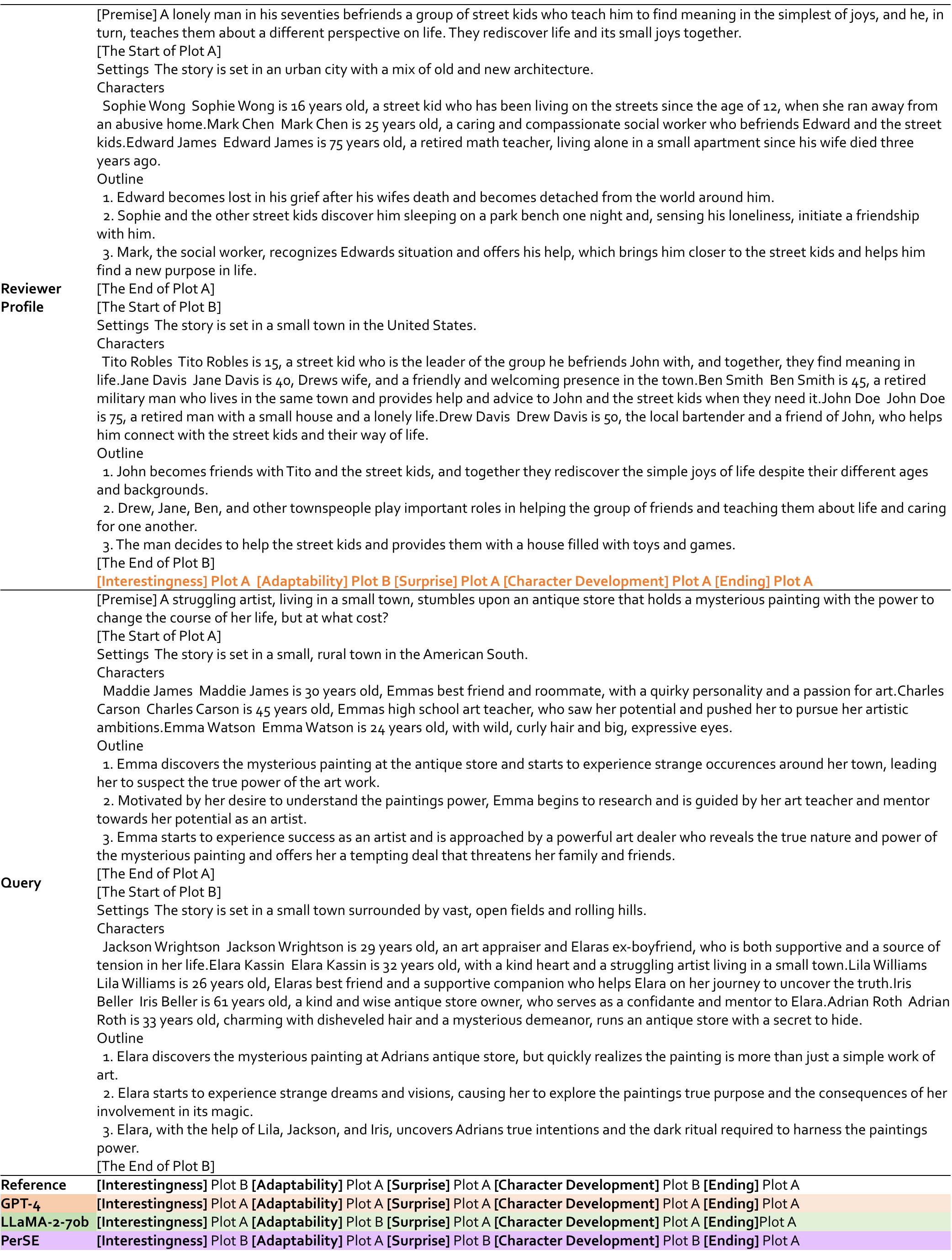}
    \caption{One case of comparative evaluation on \datasetb. \method is more similar to this reviewer. However, it fails to capture the preference of \texttt{Surprise} in this case.}
    \label{fig:case_study_5}
\end{figure*}

\textbf{\method Achieves Better Performance on Fine-Grained Pairwise Rating.} We illustrate an example from \datasetb in Figure \ref{fig:case_study_5}. \method successfully predicts the preference on 4 out of 5 aspects, while GPT-4 correctly predicts 3 aspects and LLaMA-2-70b succeeds on only 2 aspects. GPT-4 predicts Plot A for all aspects, ignoring the differences between aspects and providing an overall evaluation. In contrast, \method focuses more on the distinctive attributes of each aspect and makes judgments accordingly.

\end{document}